\documentclass[letterpaper, 10 pt, conference]{ieeeconf}  
\usepackage{graphicx}
\usepackage{amsmath, color}
\usepackage{cite, epstopdf, makeidx}
\usepackage{subfigure}

\IEEEoverridecommandlockouts                              

\title{\LARGE \bf
Evolving Unipolar Memristor Spiking Neural Networks
}

\author{David Howard, Larry Bull and Ben De Lacy Costello
\thanks{David Howard is with the CSIRO Autonomous Systems Program, QCAT, 1 Technology Court Pullenvale Brisbane 4069 Australia
        {\tt\small david.howard@csiro.au} Larry Bull and Ben De Lacy Costello are with the University of the West of England, Frenchay Campus Coldharbour Lane Bristol BS161QY UK
        }%
}

\begin{document}

\maketitle
\thispagestyle{empty}
\pagestyle{empty}

\markboth{Howard, Bull and De Lacy Costello}{Evolving Unipolar Memristor Spiking Neural Networks}

\title{Evolving Unipolar Memristor Spiking Neural Networks}
\author{David Howard$^{1,2}$, Larry Bull$^2$, and Ben De Lacy Costello$^2$\\$^1$Autonomous Systems Program, CSIRO. \\$^2$ Computer Science and Creative Technologies, \\Univeristy of the West of England.}
\maketitle

\begin{abstract}
Neuromorphic computing --- brainlike computing in hardware --- typically requires myriad CMOS spiking neurons interconnected by a dense mesh of nanoscale plastic synapses.  Memristors are frequently cited as strong synapse candidates due to their statefulness and potential for low-power implementations.  To date, plentiful research has focused on the bipolar memristor synapse, which is capable of incremental weight alterations and can provide adaptive self-organisation under a Hebbian learning scheme.  In this paper we consider the Unipolar memristor synapse ---  a device capable of non-Hebbian switching between only two states (conductive and resistive) through application of a suitable input voltage --- and discuss its suitability for neuromorphic systems.  A self-adaptive evolutionary process is used to autonomously find highly fit network configurations.  Experimentation on two robotics tasks shows that unipolar memristor networks evolve task-solving controllers faster than both bipolar memristor networks and networks containing constant nonplastic connections whilst performing at least comparably.\\
\end{abstract}

\section{Introduction}
Neuromorphic computing \cite{meadneuromorphic} is concerned with developing brainlike information processing, and requires the creation of hardware neural networks of appropriate scale together with associated learning rules.  Typically, densely-packed CMOS spiking neurons~\cite{cmos-overview} communicate with each other via voltage pulses sent along nanoscale synapses.  The memristor~\cite{chua-mem} (memory-resistor) is a two-terminal circuit element that can change between various resistance states in an analog manner through application of a suitable input voltage.  Memristors display statefulness (resistance changes are chemical in nature, so persist indefinitely and require no power to store) and a context-sensitive memory (a memristors instantaneous resistance value depends on the past of voltage activity it has experienced).  Statefulness alleviates typical nanoscale concerns regarding heat and power consumption, and context-sensitive memory allows for synapse-like information processing.  Combined, these features make memristors strong candidates for the role of synapse in neuromorphic spiking networks.

An adaptive self-organising mechanism is required to bestow learning abilities to the neuromorphic network.  Hebbian learning rules~\cite{hebb43} provide a biologically-realistic way to alter synaptic resistance values in a context-sensitive manner, depending on the activities of the neurons they connect to.  Conveniently, the adaptive resistance found in memristors closely replicates the biological plasticity observed by Hebb, and as such the two are often paired~\cite{stdp-nano-cmos-asyn, linares-barranco, stdp-discrete} to allow for adaptive learning by permitting each synapse a gradual, analog traversal over a continuous range of resistance values.

Although much current research is devoted to the use of this type of device (and this type of plasticity), there is in fact a distinction to be made between two types of memristor --- the one capable of analog, Hebbian plasticity, which is frequently called simply the ``memristor'', but which may more correctly be called the ``bipolar memristor'', and the less-discussed ``unipolar memristor''~\cite{ionics-RSM}, which shares the statefulness and memory but switches in a binary fashion between only two resistance states.  They are interesting to us as, when physical manufacture is considered, unipolar memristors are much simpler to reliably create and more durable, meaning that they are more viable candidates for physical realisation.  In this paper we simulate and analyse unipolar memristor networks, and ascertain the suitability of the unipolar memristor when used as an alternative to the bipolar memristor as a synapse in spiking neural networks.

We employ a Genetic Algorithm (GA)~\cite{holland75ga} to automatically discover high-performance spiking network topologies where each synapse is a unipolar memristor.  These networks are compared to identically-evolved benchmark networks consisting of (i) bipolar memristor synapses, and (ii) constant (nonplastic) synapses on two simulated robotics scenarios (one purely reactive, one requiring adaptation).  Results show that by foregoing the biological plausibility of bipolar plasticity, networks comprised of homogenously-parameterised unipolar memristors can adapt to dynamic environments more expediently than either of the benchmark networks without a significant degradation in other key metrics.  When coupled with the comparative ease of manufacture compared to their bipolar counterparts, unipolar memristors are highlighted as a promising, although currently overlooked, route towards the creation of physical neuromorphic architectures.

Original contributions include the introduction of such unipolar memristor networks and an analysis of the role of plasticity in the unipolar networks.  Finally, the use of two test scenarios (one reactive, one dynamic) allows us to accurately gauge the computational properties of the unipolar memristor networks in terms of behaviour generation and adaptability, both of which are key requirements for neuromorphic architectures.

\section{Memristive Spiking Networks}
\label{background}

Spiking Neural Networks (SNNs) model neural activity in the brain to varying degrees of precision.  Two well-known phenomenological implementations are the Leaky Integrate and Fire (LIF) model and the Spike Response Model (SRM)~\cite{spiking-n-m}, with the most well-known mechanistic alternative being the Hodgkin-Huxley model~\cite{hodgkin-huxley}. A SNN comprises a number of neurons connected by numerous unidirectional synapses.  Each neuron has a state, which is a measure of internal excitation, and emits a voltage spike to all forward-connected neurons if sufficiently excited.  This state is a form of memory which allows the network to solve temporal problems.

The memristor was first theoretically characterized by~\cite{chua-mem}, and first manufactured from titanium dioxide by HP labs~\cite{missing-mem-found}.  This fabrication has led to numerous other groups creating memristors from metal oxides and a variety of materials, e.g. conductive polymers~\cite{peo-0}, metal silicides, and crystalline oxides~\cite{macro-memristor}.

\begin{figure*}
\begin{minipage}{138mm}
\begin{center}
\subfigure[]{ \includegraphics[height=3cm, width=6cm]{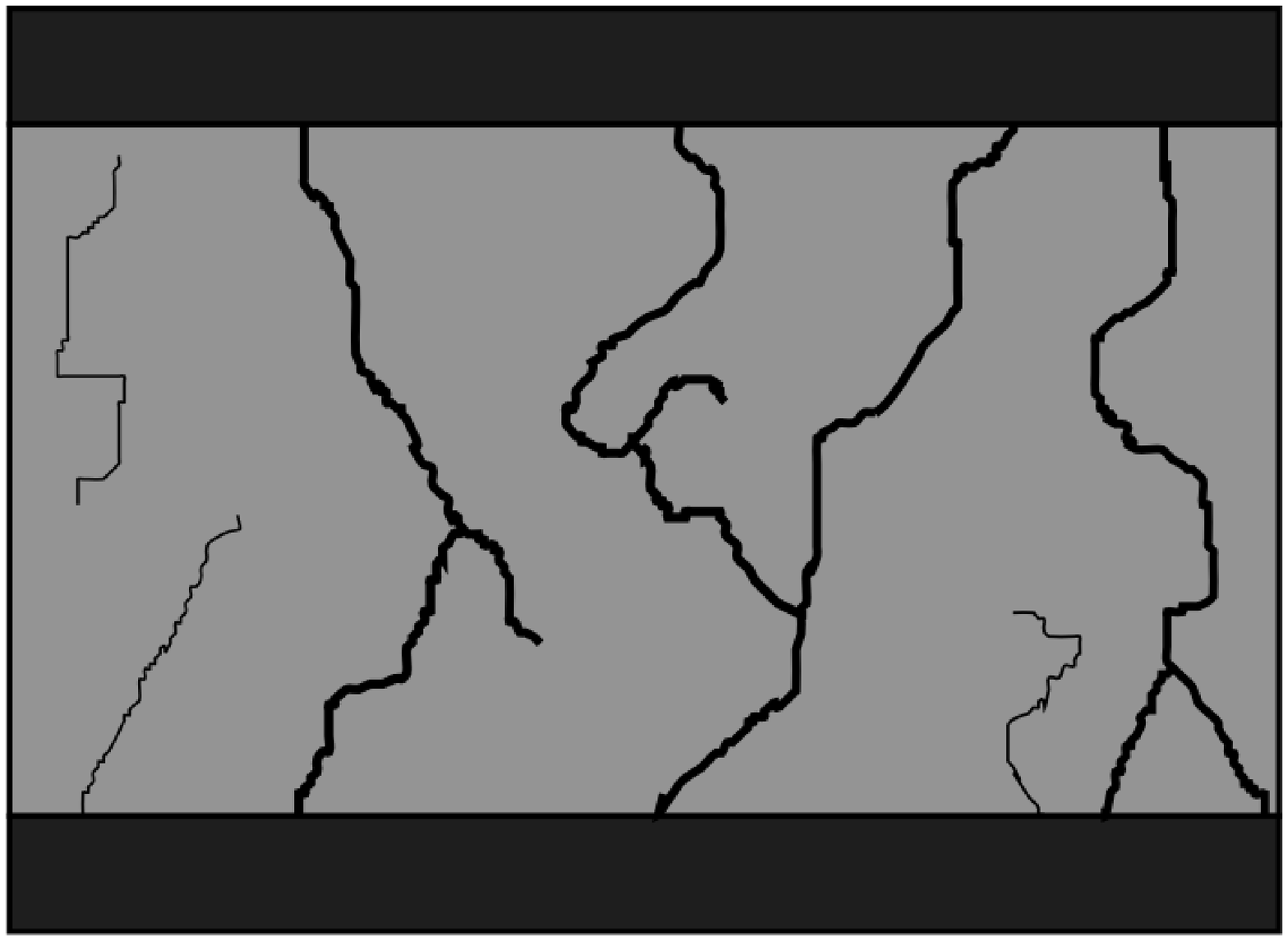}}
\subfigure[]{ \includegraphics[height=3cm, width=6cm]{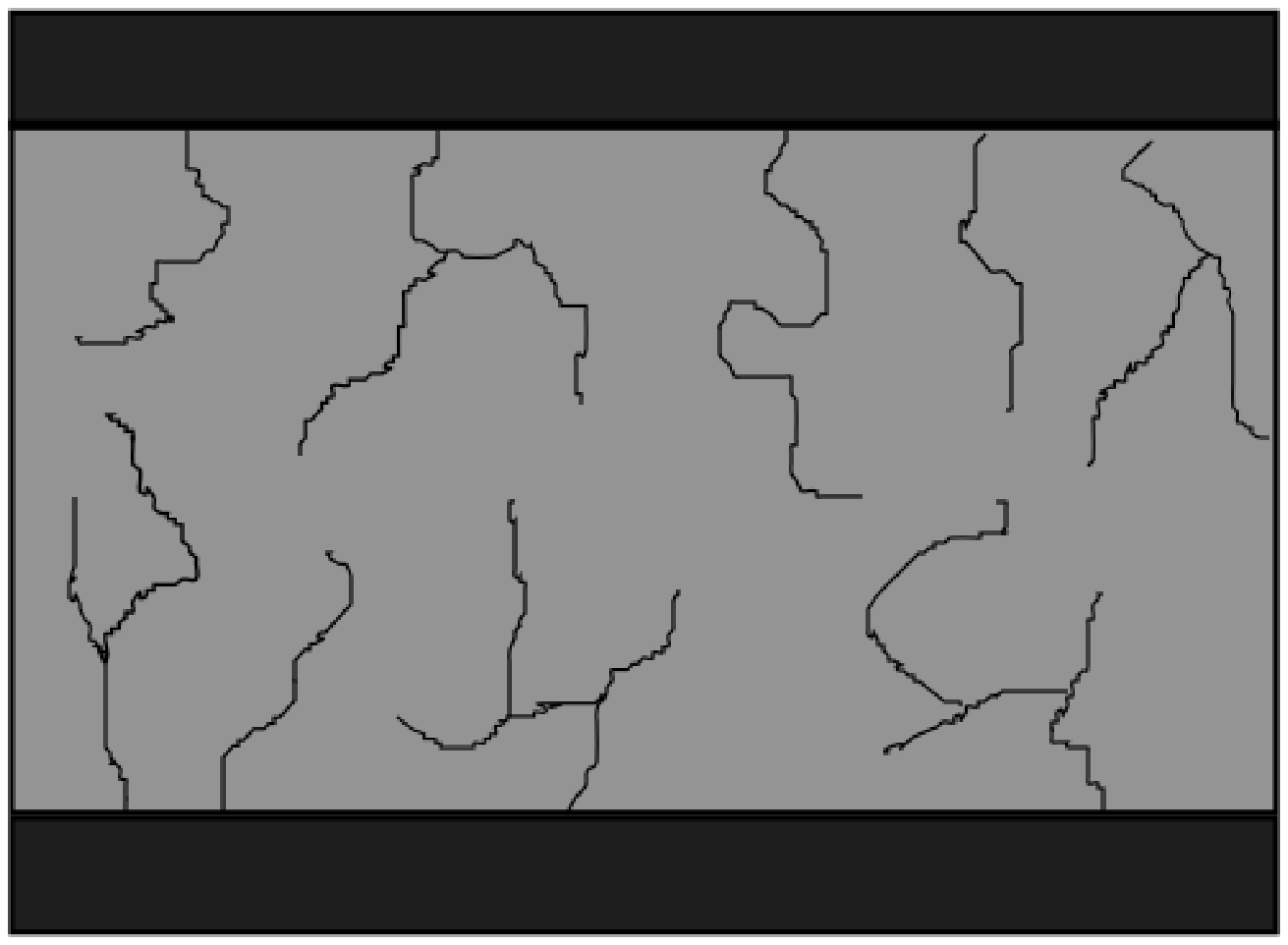}}\\
\subfigure[]{ \includegraphics[height=3cm, width=10cm]{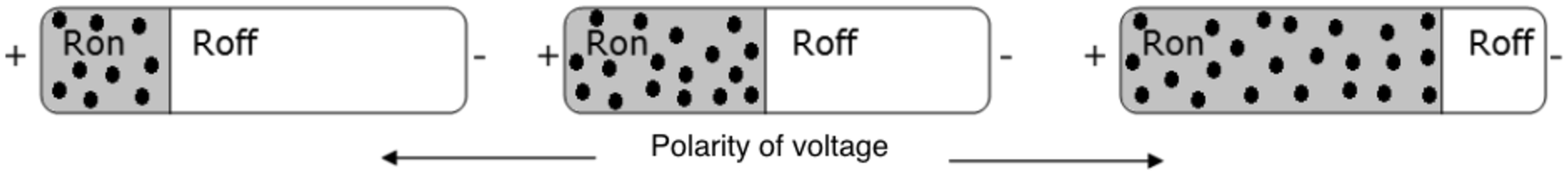}}
\caption{\label{figure1} Showing the differences in resistance change with (a, b) unipolar and (c) bipolar memristors.  Unipolar memristors form complete filaments, allowing electrons to travel largely unimpeded through the substrate and resulting in a very low resistance when formed (a), but a very high resistance when broken (b).  The bipolar memristor (which could also be represented in (b)) does not form these filaments and so relies instead on ionic conductivity (c).  The device is abstractly modelled as two variable resistors $R_{off}$ and $R_{on}$; the boundary between the resistors represents the movement of oxygen vacancies from electrode to electrode and changes as a function of applied voltage. }
\end{center}\end{minipage}
\end{figure*}

According to filament theory~\cite{ionics-RSM}, both memristor types can form internal conductive pathways called {\em filaments}, which may arise due to material defects or conditions during synthesis.  Unipolar memristors form complete filaments (Figure~\ref{figure1}(a)), resulting in drastic changes in resistance.  Mechanistically, the unipolar memristor acts as a device whose resistance can change between two values --- the Low-Resistance State (LRS) (Figure~\ref{figure1}(a)) and the High-Resistance State (HRS) (Figure~\ref{figure1}(b)) --- through application of a voltage over some threshold.  The memristor enters the LRS when complete filaments are formed.  Driving over a threshold voltage breaks these filaments and transfers the device to the HRS.  A further voltage input of suitable magnitude reforms these filaments and reinstates the device to the LRS.  Unipolar devices are ambivalent to the polarity of the applied voltage.

Bipolar memristors do not form complete filaments (Figure~\ref{figure1}(b)), meaning they must instead use comparatively weaker mechanisms such as ionic transport to alter their resistance to any value between the minimum and maximum resistance of the device in a continuous, analog manner.  The ``classic'' HP bipolar memristor can be thought of as comprising two regions, one of titanium dioxide, and the other of more conductive oxygen-depleted titanium dioxide, which are represented respectively by variable resistors $R_{off}$ and $R_{on}$.  Voltage across the device causes the oxygen vacancies to drift, altering the position of the boundary and changing the resistance depending on the polarity of the applied voltage (see Figure~\ref{figure1}(c)).  Note that the unipolar memristor also features this ionic transport; complete filament formation is simply a stronger form of resistance change so ionic effects are largely mitigated.

 Memristive plasticity involves a bidirectional voltage spike (a discrete or continuous waveform), which is emitted by a sufficiently excited neuron and can be used to track the coincidences of spikes across a synapse.  Two voltage spikes (one from each neuron that the synapse connects) arriving at the synapse in a short time window causes a coincidence event to take place.  A single coincidence event causes a bipolar memristor to change its resistance by a small amount under a Hebbian learning scheme (which when used in the context of computer science is more correctly termed Spike Time Dependent Plasticity, or STDP)~\cite{bi-poo}.  This scheme means that the order of spike arrival at a synapse affects the polarity of the voltage experienced by the device and thus the direction of synaptic weight change.  Briefly, if the presynaptic neuron fires first it can be said to have caused the postsynaptic neuron to fire and the synaptic weight is strengthened (the classic Hebb rule).  To prevent weight saturation the synaptic weight is weakened if the postsynaptic neuron fires first as such causality cannot be implied (the anti-Hebb rule).  Although we focus on unsupervised Hebbian learning, we note that supervised learning approaches also exist for memristive neural networks~\cite{Chabi}.

Unipolar memristors are less sensitive to voltage than bipolar memristors, and as such require multiple coincidence events to flip between their two resistance states. Unipolar memristors are ambivalent to the polarity of incoming voltage, so there is no notion of spike order affecting weight change and as such the scheme cannot be considered Hebbian. {In more detail, the unipolar synapse switches its resistance after a number of repeated coincidence events within a given timeframe, but is not sensitive to which neuron fires first.  A bipolar synapse requires only one coincidence event to switch, but the direction of weight change depends on which neuron (presynaptic or postsynaptic) first first, and multiple repeated coincidences in a given direction are required to affect a behavioural change in the network.  Note that unipolar plasticity removes the element of biological realism, deriving a switching mechanism more from electronic circuitry than neural circuitry, but nevertheless implements plasticity and allows for adaptivity.

In practice, the bistable nature of the unipolar memristor synapse means that the network forgoes traditional plastic means of behaviour generation and alteration. Typical bipolar plasticity involves either sensory inputs of internal neuron/synapse states are used to drive some ``trigger'' neurons into firing, which alters the plasticity of some synapses and drives the network into a different region of attractor space, potentially changing the firing rate at the output neurons~\cite{howardTEC}.  Unipolar memristors must set up weight oscillators in the network, so that coordinated synaptic weight switching leads to a suitable amount of spikes being received by the output neurons to generate an appropriate action.  Again, sensory inputs or internal network states can change the switching dynamics of the weight oscillator (e.g., the amount of time between a switch event for each synapse in the oscillator), leading to changes in behaviour.

Due to these non-standard network dynamics, the hand-design of such networks is far from intuitive.  It should also be noted that the eventual aim of this work is in automatically creating neuromorphic architectures for specific tasks, which implies some measure of self-organised learning and the freedom to discover suitable networks per-task.  Given these considerations, we have elected to use a genetic algorithm to automatically explore the space of network topologies.  Related work on neuro-evolution includes~\cite{neuroevo-arch-learn}, who survey various methods for evolving both weights and architectures.~\cite{evo-rob-book} describe the evolution of networks for robotics tasks.  Combined with plasticity, neuro-evolutionary controllers have shown increased performance compared to similarly-evolved nonplastic-synapse networks~\cite{soltoggio08, flor-urz}. Related research on neuromorphic memristive networks includes the use of conductive polymers~\cite{ero-neuro} and crossbar implementations ~\cite{Kim2015}.

Numerous groups have previously used bipolar memristors as plastic synapses \cite{stdp-nano-cmos-asyn, linares-barranco,stdp-discrete}, seeking to exploit the similarities between analog resistance alteration and Hebbian learning (e.g.,~\cite{7086091}).  To our knowledge, no previous work has considered the use of unipolar memristor synapses to fulfil the same role.  Examples of unipolar memristors are confined to binary operation for traditional logic~\cite{unipolar-mem}, although neural implementations exist for non-memristor binary switching devices~\cite{guangpu}.  Although we note that spiking unipolar memristor networks have not been studied in detail before, similar studies give cause for optimism.  A string of publications from the group of Stefano Fusi focuses on the how the number of available synaptic resistance states affects memory and learning in neural networks e.g.,~\cite{Fusi2005599,fusi2007limits}.  In particular, they conclude that binary synapses (similar to our unipolar memristors) allow for learning given certain prerequisites are met~\cite{1243728}.  \cite{soltoggio2012modulated} uses synaptic weight saturation that is sensitive to initially-noisy synaptic states to a similar effect.

\subsection{Motivation for the Focus on Unipolar Memristors}
We focus on unipolar memristors because they are easier to implement in reality than bipolar memristors --- using binary rather than analog resistance states means that the devices are less reliant on precise nanoscale fabrication, which can be compromised by device variations that are currently an intrinsic part of the nanoscale manufacturing process.  In other words, the device only needs to reliably switch between two resistance states, rather than follow a specific resistance profile to a given degree of accuracy. This benefit extends to the testing of physical networks --- ``does it switch?'' is an easier question to answer than ``does it follow this continuous resistance profile accurately enough?'', and requires less time to test.  Similar benefits have been reported by \cite{1243728} --- reducing the amount of device variance is highlighted as a route to a simpler manufacturing process, e.g. in the context of Very Large-Scale Integration (VLSI).

During operation, unipolar memristors require multiple spike conincidence events to force a single state change; it follows that they undergo fewer total state changes and therefore may be more long-lasting.  A single resistance change can perturb the network state more than is possible in bipolar networks, potentially leading to simpler implementations requiring reduced spike traffic (and hence lower power requirements).  Simpler network activity through a lower-dimensional attractor space may permit a more tractable analysis and understanding of candidate networks.  In summary, they are simpler to reliably and repeatedly create, and may give rise to lower-power and simpler implementations than bipolar memristors.

\section{Tasks}

Two robotics tasks are used to evaluate the unipolar synapse networks in the context of the other network types.  The first --- phototaxis --- is purely reactive, and designed to illustrate the quality of behaviour that can be elicited from each network type.  The second task, performed in a T-maze, shows the speed of adaptation to a dynamically-changing environment.

\begin{figure*}
\begin{center}
\subfigure[]{ \includegraphics[width=6cm,height=6cm]{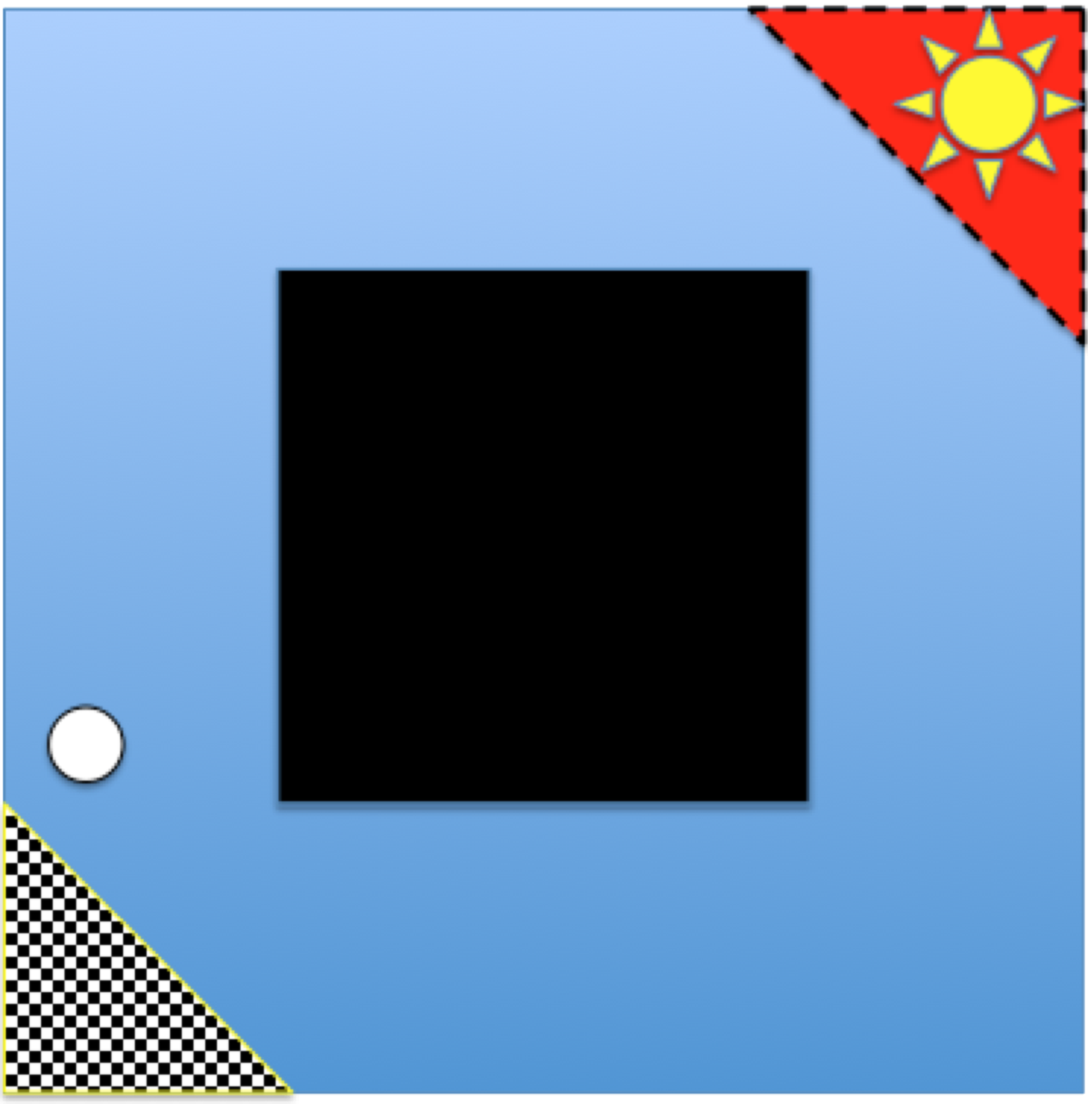}}
\subfigure[]{ \includegraphics[width=6cm ,height=6cm]{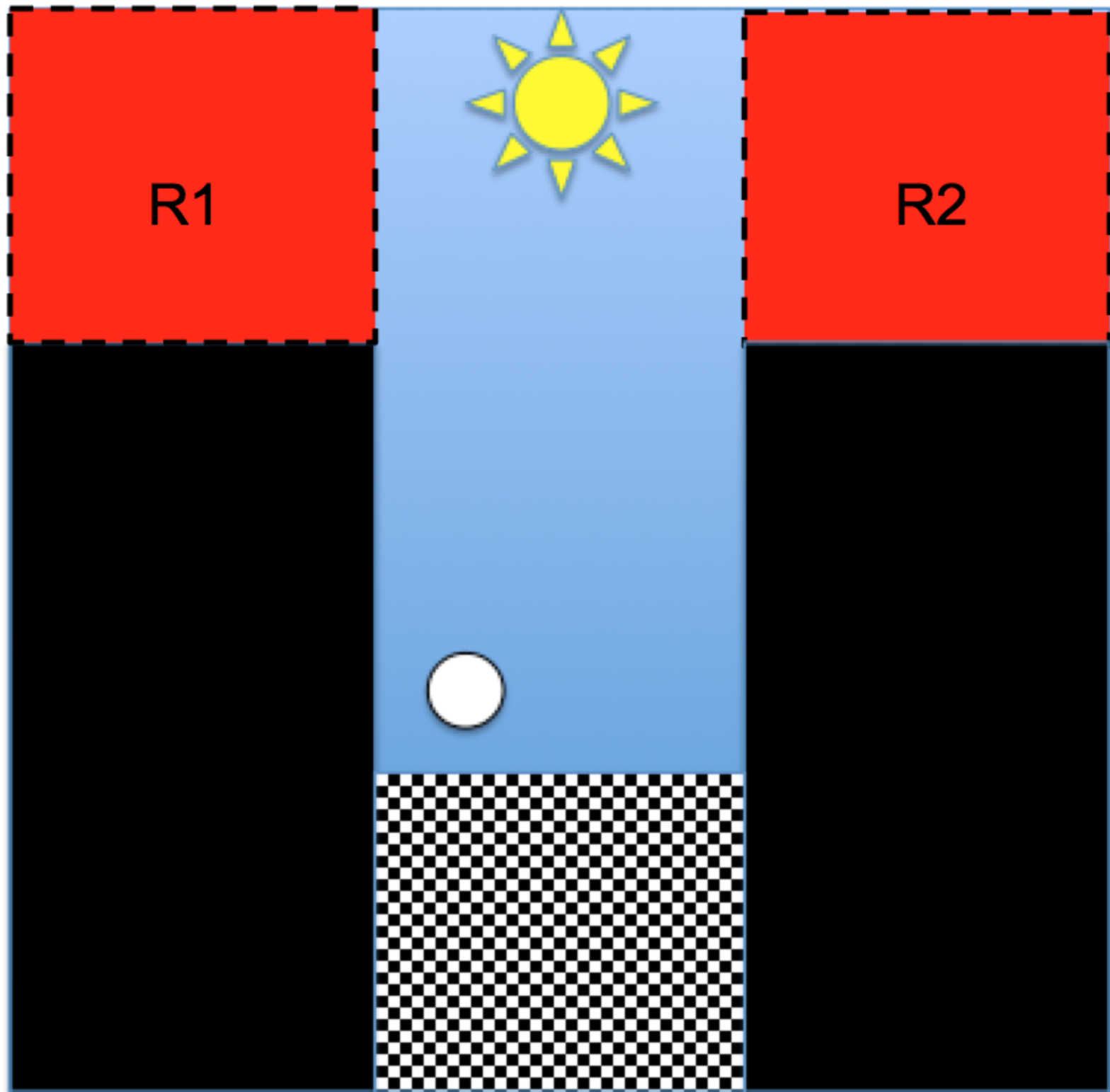}}
\caption{\label{figure2} (i) The phototaxis environment. The agent (circle) begins randomly positioned in the lower-left (checkered triangle) and must reach a light source in the upper-right, circumnavigating the central obstacle. (ii) The T-maze.  The robot (circle) begins randomly positioned in the start zone (checkered box) and must travel to reward zones R1 or R2.  A light source is located in the top-centre.  Both figures are to scale.}
\end{center}
\end{figure*}

\subsection{Task I: Phototaxis}
The phototaxis task is classed as reactive as it does not require memory or adaptation to solve.  In the phototaxis environment, a differential drive robot is randomly located within a walled arena, with boundaries at -1 and 1 in both $x$ and $y$.  A three-dimensional box, which the robot must navigate around, is placed centrally in the arena, with vertices at ($x=-0.4$, $y=-0.4$), ($-0.4$, $0.4$), ($0.4$, $0.4$), and ($0.4$, $-0.4$).  A 15 Watt bulb is placed at the top-right hand corner of the arena ($x=1$, $y=1$).  The environment is shown in Figure~\ref{figure2}(a).

The robot initially faces North, with an initial start position randomly generated but constrained $x+y < -1.5$.  The robot must perform phototaxis --- light-seeking behaviour --- and receives fitness proportional to the closest distance it achieves to the light source.  When the robot reaches the goal state (where $x+y>1.6$), the responsible controller receives a constant fitness bonus of 2500, which is added to the fitness function $f$ ($f>0$) outlined in equation (~\ref{photo_fit}).  The denominator in the equation expresses the difference between the position of the goal state (1.6) and the current robot position ($x_{pos}$ and $y_{pos}$), and $st$ is the number of robot steps taken to reach the goal state.  A simple step-based fitness could have been used, but our fitness function allows for a gradual improvement of behaviour that does not require the goal state to be found to begin optimisation.

The fitness of a controller is calculated at the end of every robot step, with the highest attained value of $f$ during the trial kept as the fitness value for that controller.  Optimal performance gives $f=11800$, which corresponds to 700 robot steps from start to goal state with no collisions.

\begin{equation}
f = (1/(1.6 - (|x_{pos}+y_{pos}|))) \times 1000 - st
\label{photo_fit}
\end{equation}

\subsection{Task II: Adaptation}

A dynamic T-maze~\cite{tmaze,solt-dyn} scenario is used to measure the adaptation capabilities of the synapses.  The T-maze is an enclosed arena with coordinates ranging from [-1,1] in both $x$ and $y$ directions, with walls placed to represent a ``T'' (Figure~\ref{figure2}(b)).  Reward zones R1 and R2 are situated at the end of the left and right arms respectively.  A 15 Watt bulb is placed at the top-centre of the arena ($x=0.5$, $y=1$) and is used to indirectly feed position information to the network, as well as enabling it to produce any action from anywhere in the arena.

At the start of a trial, a differential-drive robot initially faces North, randomly positioned in the start zone at the bottom of the ``T'' ($-0.4 > x < 0.4$, $y<-0.4$).  The agent must navigate to the initial reward zone R1.  The trial is split into two phases, each of which is 4000 robot steps long for a maximum trial length of 8000 steps.  Phase 1, similarly to the first task, evolves controllers that can navigate from the start zone to R1.  Arriving in R1 resets the robot in the start zone and commences phase 2.  Any controller that reaches R1 is immediately retested 5 times to ensure that the pathfinding is stable.

In phase 2, the adaptation of the network is measured by switching the reward zone to R2.  To give the robot memory of phase 1, membrane potentials and synaptic weights are not reset during this process.  By measuring the number of generations that each network type takes to adjust to the new goal position, we quantify how quickly the network can adapt it's behaviour to match the dynamic reward zone.  Again, 5 retests are carried out to ascertain the stability of the solution.

The aim is to measure the length of adaptation, encapsulated in the ``solved'' generation of the network.  The fitness function, $f$, is simply the total number of robot steps required to solve the trial (equation (\ref{tmaze_fit}) --- lower fitness is better).  A controller that cannot locate R1 receives maximum fitness (8000) for the trial.  A controller that locates R1 but cannot subsequently adapt to R2 receives maximum fitness for R2 (4000), plus however many robot steps it took to locate R1.  Fitness measures quality of pathfinding, and ensures that the best networks optimise towards useful goal-seeking behaviour.

\begin{equation}
f = st
\label{tmaze_fit}
\end{equation}

\subsection{Experimental Procedure}

At the start of each experiment, 100 spiking networks are randomly generated.  The synapse type used varies per experiment, being either a unipolar memristor, a bipolar memristor, or a constant connection.  Each network is then evaluated on the task over a maximum of 8000 robot steps.  Each robot step involves the network processing its sensory input for a number of processing steps, after which the spike trains generated at the output nodes are used to select an action.  The robot executes the action, concluding the robot step, and recieves the next sensory input as the first part of the subsequent robot step.  After a trial (which ends either with success or timeout), each controller is assigned a fitness.  A genetic algorithm then optimises the population of networks for a task-dependent number of generations.

\section{Spiking Controllers}
Leaky Integrate and Fire~\cite{spiking-n-m} networks are used as spiking controllers.  Three layers of neurons (input, hidden and output) have sizes of 6, 9 and 2 respectively.  On network creation, the hidden layer is populated with 9 hidden layer neurons, whose types are intitially excitatory (transmit voltages $V$$\geq$ 0) with P=0.5, otherwise they are inhibitory ($V$$<$0).  Each connection has a weight $w$ (all weights constrained [0-1]).  Each possible connection site is initially likely to have a connection with P=0.5.

During activation, stimulation by incoming voltage alters a neurons internal state $m$, $m>0$, which by default decreases over time.  Surpassing a threshold $m_\theta$ causes a spike to be transmitted to all postsynaptic neurons. The amount of voltage sent is equal to the weight of the connection, multiplied by -1 if sent from an inhibitory neuron.   The state of a neuron at processing step {\em t+1} is given in equation~\ref{eq3}; equation~\ref{eq4} shows the reset formula.  $m(t)$ is the neuron state at processing step $t$, $I$ is the scaled state input, $a$ is an excitation constant and $b$ is the leak constant.  Immediately after spiking, the neuron resets its state to $c$ following equation~\ref{eq4}.  A spike sent between two hidden layer neurons is received $n$ ($n>0$) processing steps after it is sent, where $n$ is the number of neurons spatially between the sending neuron and receiving neuron in the layer.  This implements a weak form of spatial ordering to the networks, without explicitly placing the networks on a virtual substrate.  Parameters are $a=0.3$, $b=0.05$, $c=0.0$, $m_\theta = 0.6$

 \begin{equation}
m(t+1) = m(t) + (I+a-bm(t))
\label{eq3}
\end{equation}
\begin{equation}
\mathrm{If}\; (m(t) > m_\theta)\;\;\;\; m(t) = c
\label{eq4}
\end{equation}

Our bidirectional voltage spikes are discrete-time stepwise waveforms, matching the discrete-time operation of the SNNs.  Each neuron in the network is augmented with a ``last spike time'' variable $LS$, which represents voltage buildup at the synapse and is initially 0.  When a neuron spikes, this value is set to an experimentally-determined positive number, in this case 3.  At the end of each of the 21 procesing steps that make up a single robot step, each memristor synapse is analysed by summating the $LS$ values of its presynaptic and postsynaptic neurons --- any value greater than a threshold $\theta_{LS}$=4 is said to have caused a coincidence event at the synapse.  Each $LS$ value is then decreased by 1 to a minimum of 0, creating a discrete stepwise waveform through time, see Figure~\ref{figure4}(a).\\

\subsection{Controller Integration}
Both experiments use the same differential drive robot with 3 active light sensors and 3 active distance sensors shown at positions 0, 2, and 5 in Figure~\ref{figure3}(b).  Random-uniform sensory noise in included -- $\pm$2\% for IR sensors and $\pm$10\% for light sensors.  To prevent the robot from becoming stuck in the environment, two bump sensors are used (see Figure~\ref{figure3}(b) for placement) --- activating either causes the robot to immediately reverse 10cm (an effective penalty of 10 robot steps spent reversing).\\

\begin{figure*}
\begin{minipage}{138mm}
\begin{center}
\subfigure[]{ \includegraphics[width=6.5cm,height=6.5cm]{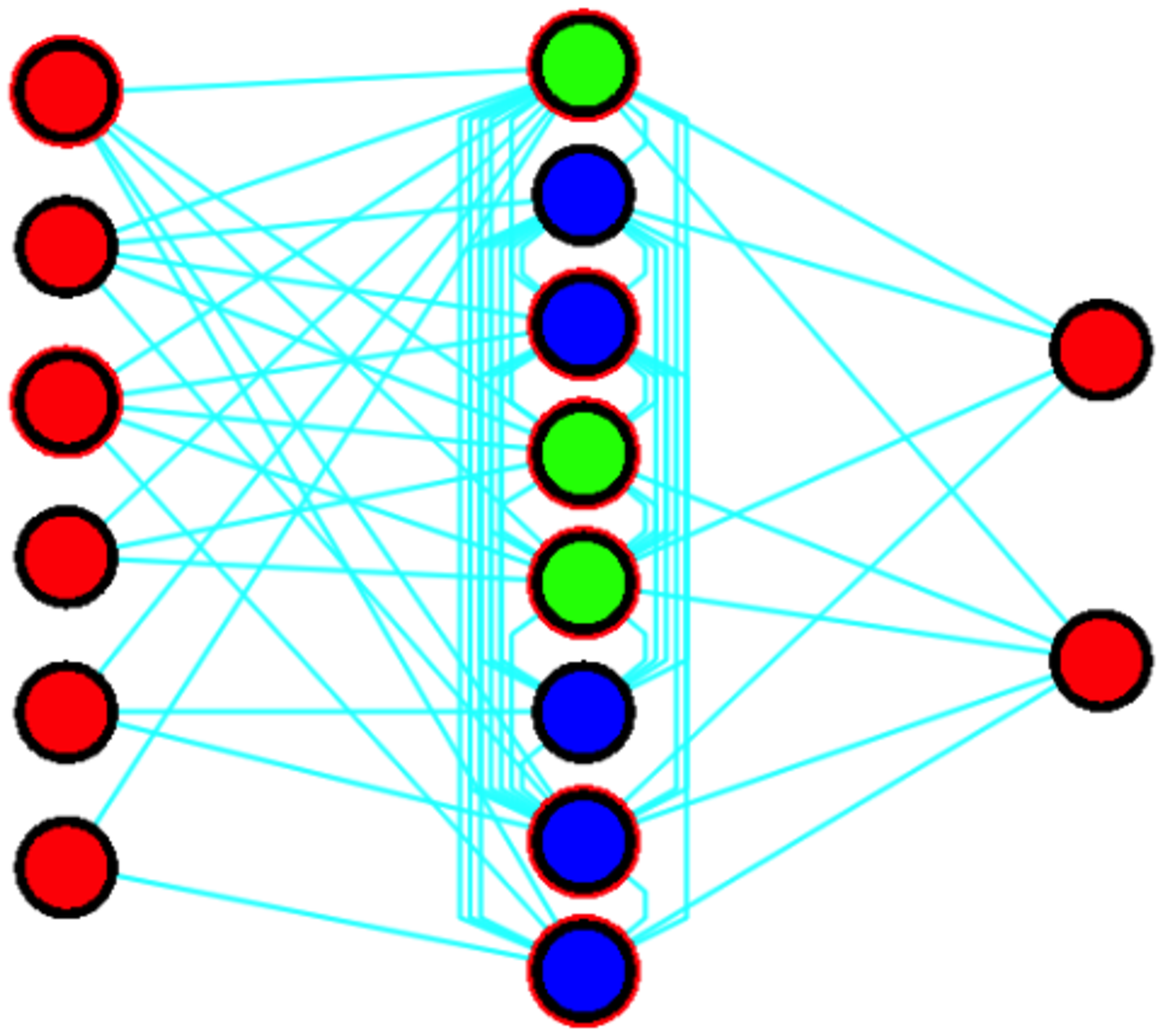}}
\subfigure[]{ \includegraphics[height=6.5cm, width=6.5cm]{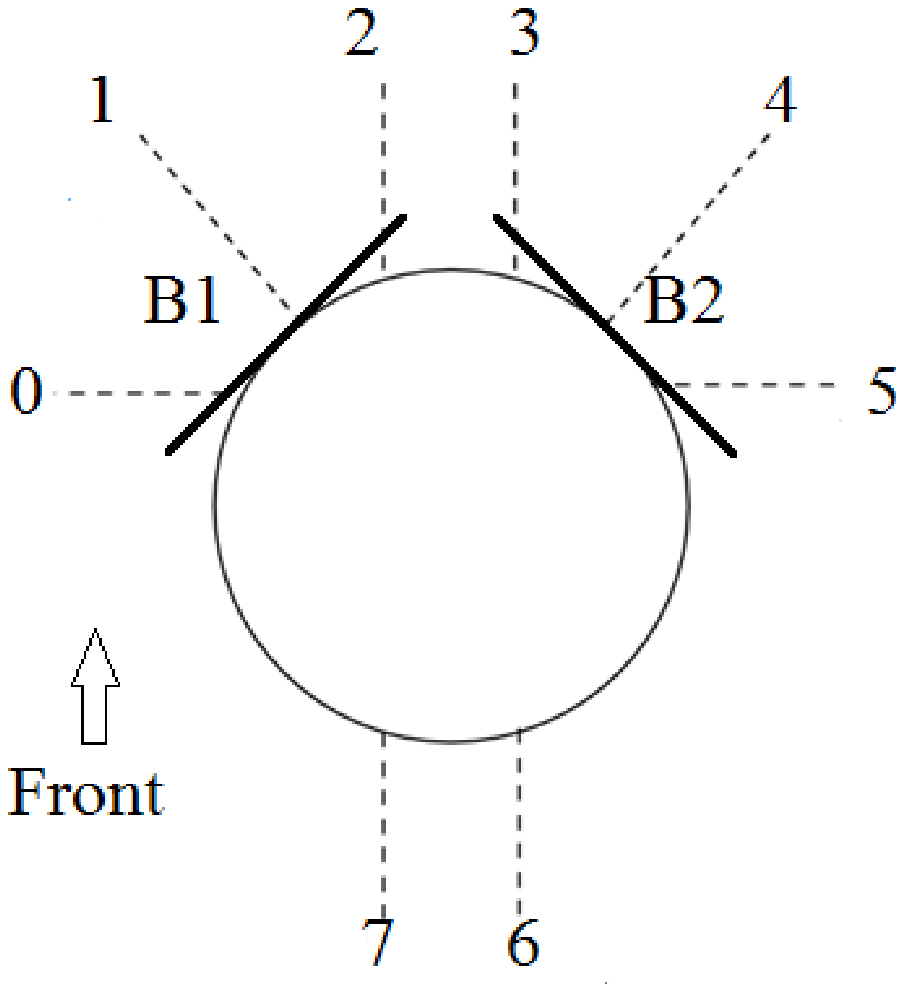}}
\end{center}
\caption[]{\label{figure3} (a) A typical spiking network architecture.  The top 3 input neurons receive light sensor activations, the bottom 3 receive IR sensor activiations, and spike trains generated at the two output neurons are used for action calculation.  In the hidden layer, green/light neurons denote excitatory neurons and blue/dark neurons signify inhibitory neurons. (b) The differential drive robot used in both experiments.  There are 8 sensor mounting positions (labelled 0-7).  In our setup, three light sensors and 3 IR sensors share positions 0, 2, and 5 and form the network input.  The other numbers show unused sensor positions.  Two bump sensors, B1 and B2, are shown attached at 45 degree angles to the front-left and front-right of the robot.}
\end{minipage}
\end{figure*}

 For each robot step (64ms in simulation time), the robot samples the six sensors: the six-dimensional input vector is then scaled so that the entire sensor range falls within [0,1], and is used as $I$ in equation (\ref{eq4}). The network is then run for 21 processing steps --- experimentally determined to allow the bipolar synapses enough time to change synaptic plasticity to affect useful behaviour generation --- and the spike trains at the output neurons discretised as having either {\em high} or {\em low} activation to generate an action ({\em high} activation if more than half of the 21 processing steps generated a spike at the neuron,  {\em low} otherwise).  Three discrete actions are used to enforce more distinct changes in network activity and to encourage more differentiation between the activity of the different synapse types.  Actions are {\em forward}, (maximum movement on both wheels, {\em high} activation of both output neurons) and continuous turns to both the {\em left} ({\em high} activation on the first output neuron, {\em low} on the second) and {\em right} ({\em low} activation on the first output neuron, {\em high} on the second) --- caused by halving the left/right motor outputs respectively.

\subsection{Synapse Types}
In this section we describe our implementation of the three synapse types used in the experiment.  Both memristor synapses rely on the concept of a ``coincidence event'', which is defined as two spikes arriving at the synapse at consecutive processing steps.  The {\bf unipolar memristor synapse} has parameters $S_n$, which represents the sensitivity of the device to voltage buildup (in the form of repeated coincidence events), and $S_c$, which tracks the number of consecutive coincidence events the synapse has experienced.  All synapses are initially in the Low Resistance State ($w=0.9$).  This is an arbitrary selection, performance is unaffected if the networks begin in the HRS.  Starting in the HRS would just start the network in a different stage of its weight oscillator, which is initially determined by the initial network topology, whilst preserving the coincidences of weight switches which are required to generate the output action.

At each processing step every synapse is checked, incrementing $S_c$ if a coincidence event occurs at the synapse and decrementing $S_c$ if no coincidence event occurs at that processing step.  If $S_c$=$S_n$ (Figure~\ref{figure4}(a)), the unipolar memristor switches to the HRS ($w=0.1$) if it was previously in the LRS, or the LRS if it was previously in the HRS.  $S_c$ is reset to 0.  The device can switch between these states multiple times per trial.  Due to the requirement of multiple consecutive coincidence events per switch, the actual frequency of synapse alteration is lower than that seen in the bipolar networks.  Note that switching between two resistance states, maximally resistive and minimally resistive, likens the unipolar plasticity mechanism to network-wide feature selection, rather than online weight adaptation as with traditional Hebbian plasticity schemes.

Initial experimentation (excluded for the sake of brevity) performed a sensitivity analysis on the $S_n$ parameter --- no statistically significant differences were observed between values of 2, 4, and 6.  In this work we select $S_n$=4 as a compromise between switching speed and potential device longetivity in hardware implementations.\\

\begin{figure*}
\begin{minipage}{138mm}
\begin{center}
\subfigure[]{ \includegraphics[height=5cm, width=7cm]{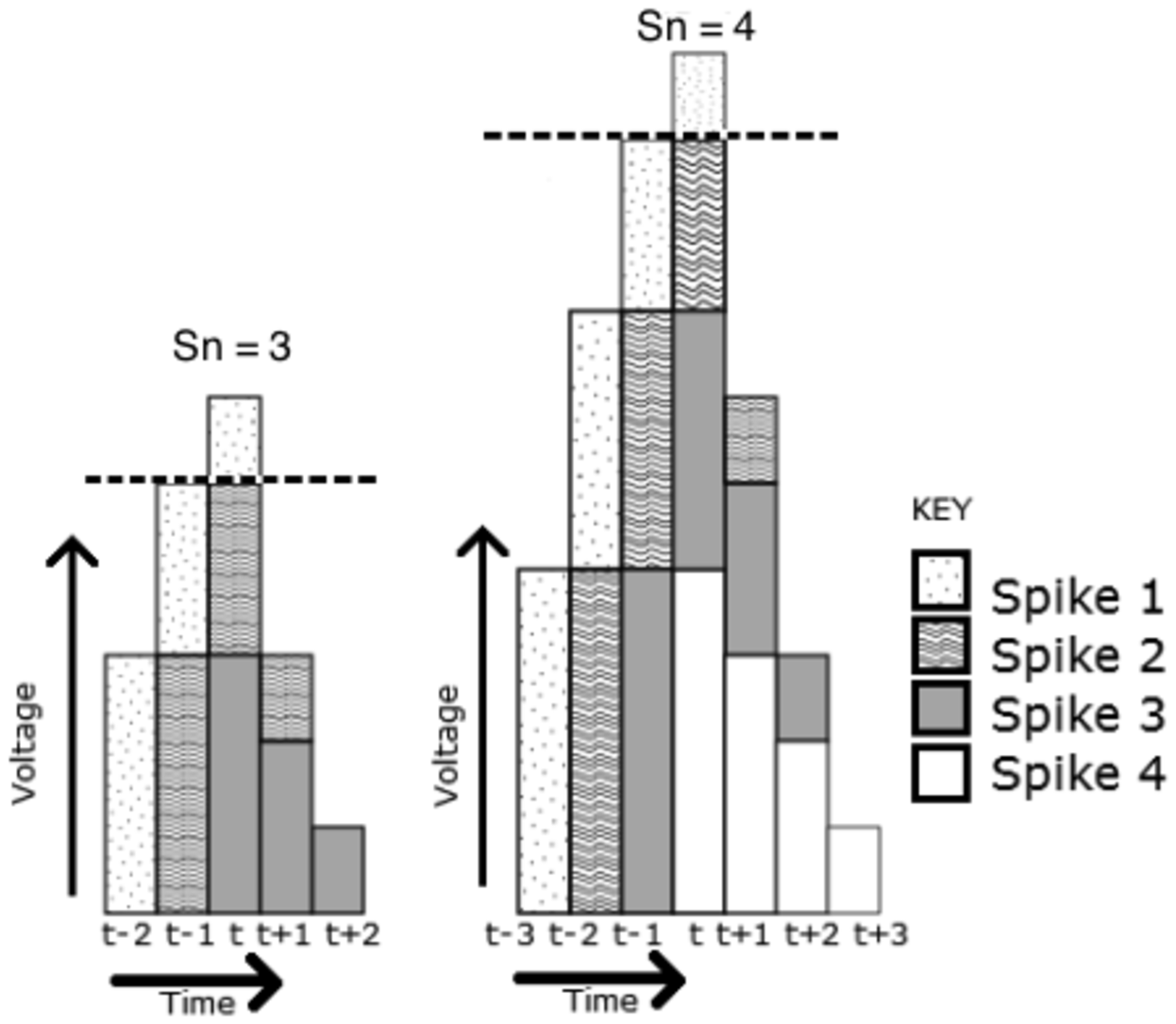}}
\subfigure[]{ \includegraphics[height=5cm, width=6cm]{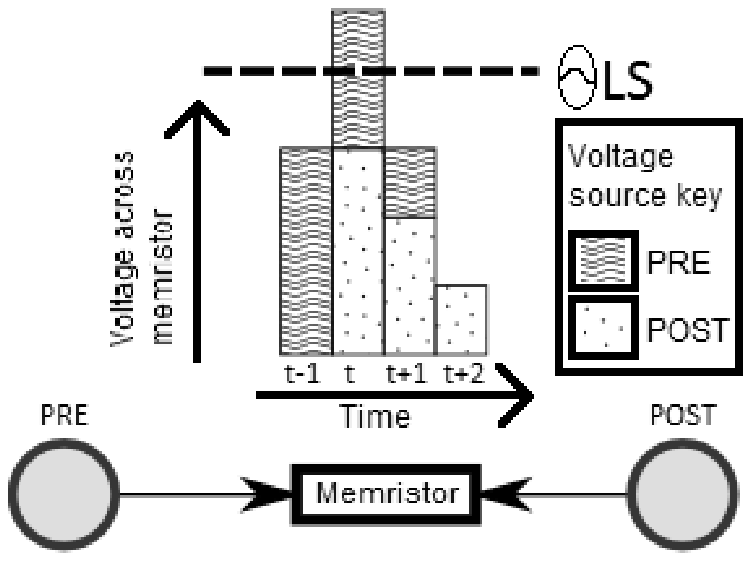}}
\end{center}
\caption{\label{figure4} Showing a spike coincidence event for both memristor types.  A presynaptic voltage spike is received at processing step {\em t-1}, with a postsynaptic voltage spike at processing step {\em t} (a ``coincidence event''). For the unipolar memristor (a) multiple subsequent events (l.h.s $S_n=3$, r.h.s $S_n=4$) are required to push the voltage past a threshold, causing a switch.  Dotted lines show the derived voltage threshold.  Voltage spike values are decremented by one per subsequent processing step. (b) A single event allows the voltage to surpass $\theta_{LS}$, decresing the resistance of the bipolar memristor by a smaller amount.}
\end{minipage}
\end{figure*}

We use a generalised model of the {\bf bipolar memristor synapse} --- as previously noted, memristor materials and fabrication techniques are highly varied, and similar variance is seen in the plastic behaviour of the synapses.  We elect to use a linear model as this provides the most bias-free comparison, whilst still providing incremental resistance changes.  This model is competitive with real-world memristor models~\cite{howardTEC}.  Bipolar memristor connection weights are initially 0.5.  Bipolar memristors have an effective $S_n$ of 1 (Figure~\ref{figure4}(a)), so every coincidence event causes a change in synaptic weight by $\Delta_w$=0.001, meaning 1000 consecutive Hebbian events will take the synapse from maximally resistive to minimally resistive.  Weight increases if the postsynaptic neuron has the highest $LS$ value, and decreases if the presynaptic $LS$ is higher, in other words the bipolar memristor is sensitive to the polarity of applied voltage.

The {\bf constant} synapse is non-plastic (essentially a resistor).  The resistance of the connection is initialised random-uniformly in the range [0,1] and may be altered during application of the GA, but is constant during a trial.  The constant connection is used as a baseline that shows the effects of having no plasticity in the network.  It should be noted that all networks have memory (in the form of neuron membrane potentials), but only the plastic unipolar and bipolar synapse networks have the additional freedom to adapt their connection weights online.

\section{Genetic Algorithm}
\label{discovery}
In our steady-state GA, two child networks are created per generation.  Two parents are selected via fitness-proportionate selection on the 100 population networks, and their genomes copied to two child networks and probabilistically mutated.  Crossover is omitted --- sufficient network space exploration is obtained via a combination of weight and topology mutations; a view that is reinforced in the literature~\cite{mu-only-needed}.  The networks are then trialled on the test problem and assigned a fitness before being added to the population.  Finally, the two worst-fitness networks are deleted from the population.  Each network has its own self-adaptive mutation rates, which are initially seeded uniform-randomly in the range [0,0.5] and mutated as with an Evolution Strategy~\cite{rechenberg} as they are passed from parent to child following equation~(\ref{eq-sa}).

\begin{equation}
\label{eq-sa}
\mu \rightarrow \mu\; exp^{N(0,1)}
\end{equation}

This approach is adopted as it is envisaged that efficient search of weights and neurons will require different rates, e.g., adding a neuron is likely to impact a network more than changing a connection weight, so less neuron addition events than connection weight change events are likely to be desirable.  Self-adaptation is particularly relevant for the application area of neuromorphic computing --- brainlike systems must be able to autonomously adapt to a changing environment and adjust their learning rates accordingly.

The genome of each network comprises a variable-length vector of connections and a variable-length vector of neurons, plus a number of mutation rates.  Different parameters govern the mutation rates of connection weights ($\mu$), connection addition/removal ($\tau$), and neuron addition/removal ($\omega$).  For each comparison to one of these rates a uniform-random number is generated; if it is lower than the rate, the variable is said to be {\em satisfied} at that allelle.  During GA application, for each constant connection, satisfaction of $\mu$ alters the weight by $\pm$0-0.1.   Memristive synapses cannot be mutated from their initial weights of 0.9 for unipolar and 0.5 for bipolar, forcing those networks to use plasticity to perform well.  Each possible connection site in the network is traversed and, on satisfaction of $\tau$, either a new connection is added if the site is vacant, or the pre-existing connection at that site is removed.  $\omega$ is checked once, and equiprobably adds or removes a neuron from the hidden layer (inserted at a random position) if satisfied.  New neurons are randomly assigned a type, and each connection site on a new neuron has a 50\% chance of having a connection. New constant connections are randomly weighted between 0 and 1.

\section{Experimentation}

We test the three synapse types (unipolar, bipolar, constant) on each environment for 30 experimental repeats, using the averages to create the statistical analysis given below.  Each phototaxis experiment is run for 1000 generations, and each T-maze experiment for 500 generations.  Two-tailed T-tests are used to asses statistical significance, with significance at P$<$0.05.  As well as fitness, we also track the first generation in which each population produces a controller that solves the problem, which we term ``success''.

The two environments test different aspects of the computational abilities of the synapses.  In both environments, fitness represents the quality of the pathfinding behaviour.  In the phototaxis task, ``success'' straightforwardly measures the number of generations to generate a controller that sucessfully reaches the goal state.  In the T-maze, ``success'' measures the speed at which the controllers can alter their behaviours in response to the change in goal state position, an indicator of adaptivity.

In both cases, we wish to find differences that strengthen our position, i.e.,  that the unipolar memristors are a viable alternative to bipolar memristors.   These tasks allow us to answer important questions regarding the power of the unipolar plasticity mechanism --- does the plasticity permit sufficient adaptivity to solve the problem? Does the discontinuous switching behaviour allow the evolutionary process to set up useful attractors more expediently?  What are the benefits and drawbacks in terms of controller performance and network composition?  In the case of the Tmaze, is the unipolar networks memory ability impacted by the binary nature of the synapse?

\subsection{Results I: Phototaxis}

Unipolar memristor networks are shown to have better final best fitness than the other two network types (Table~\ref{table1}). Figure~\ref{figure5}(a) shows that unipolar networks have a higher initial fitness ($\approx$10000) then the comparative network types (constant $\approx$4000, bipolar $\approx$8500), indicating some passable unipolar controllers in the initial population.  This is likely because of the ``constrained flexibility'' afforded to unipolar networks --- plastic online behaviour but a relatively simple attractor space that has vastly fewer dimensions than that of the bipolar networks.   This result highlights the role of switching plasticity in generating high quality pathfinding behaviour in terms of being able to generate heterogenous action sequences (a single switch can perturb network output enough to change the action, and the switching nature of unipolar plasticity allows the synapse to repeatedly switch on and off to generate heterogenous action sequences).  Following a period of low-performing controllers in generations $\approx$0-400, constant connection networks undergo a rapid fitness improvement from generations $\approx$400-500.  A stable plateau occurs for all network types at $\approx$500 generations.

Both plastic networks have the ability to search the behaviour space online as well as offline, which is reflected in their ``solved'' generation values being statistically superior to constant networks (Table~\ref{table1}).  As the constant networks must search entirely offline via the GA, they take longer to develop the required behaviour (avg. 77.6 generations to solve, compared to 0.76 for unipolar and 14.7 for bipolar).  Additionally, the unipolar networks solve statistically faster than the bipolar networks.  In concordance with the analysis of Figures~\ref{figure5}(a)-(c) above, these results show that although plasticity is beneficial in general to the evolutionary process, the more gradual Hebbian plasticity used by the bipolar networks results in a larger search space than that of the unipolar networks, resulting in slower convergence.

This notion is echoed in results for average fitness (Table~\ref{table1}).  Constant networks have statistically better final average fitness than bipolar networks --- bipolar synapses have a more complex attractor space which leads to more fitness variance in the final population as full convergence is not achieved within the generation limit.  Although the population of constant networks initially struggles to achieve uniformly high fitness values due to a lack of behavioural flexibility in the networks (large standard error and low mean in Figure~\ref{figure5}(b) between generations 200 and 600), it eventually converges due to having a simpler search space.  Unipolar and bipolar synapses are seen to approximately equal each other, with much lower standard error then the constant connection between generations $\approx$200-600.  We note that the final fitness order of the synapses (constant, unipolar, bipolar) is also the synapse complexity order, indicating that population convergence is related to the compexity of the network behaviour space that the GA has to optimise in.

 Average connected hidden layer nodes (Figure~\ref{figure5}(c)) do not vary significantly between the network types.  We note that the number of neurons in the constant networks jumps from $\approx$16.9 to $\approx$17.1 during generations $\approx$400-500, with a corresponding increase in average best fitness over the same period (Figure~\ref{figure5}(a)).  Connectivity (Figure~\ref{figure5} does not vary significantly between network types, with bipolar networks in particular displaying a large amount of standard error throughout the generations.  As the synapse itself is more variable, larger subset of connectivity maps (with varying amounts of connections) can provide the same types of behaviour.

\begin{table*}
 \caption{Phototaxis averages and standard deviations for controller parameters for the three synapse types. Symbols indicate the value is statistically (p$<$0.05) better than $^{o}$ = Unipolar $\dagger$ = Bipolar, * = Constant.}
 \begin{tabular}{@{}lcccccc}
 				& Best fit       			&  Avg fit     			& Gens. to solve    		& Nodes        &Connectivity     \\ 
Unipolar		&11718 $\dagger$*(186)		&11362 (452)			& 0.26 $\dagger$*(1.28)		&16.9 (0.4) 	&51.75 (4.47)\\ 
Bipolar			&11363  (398)				&11058 (728) 			& 14.7 *(32.5)				&16.89(0.54)	& 51.06 (4.06)   \\ 
Constant		&11420 (423)				&11402 $\dagger$ (277)	& 77.6 (130.0)				&17.10 (0.7)	&51.24 (3.58)\\ 
\end{tabular}
\label{table1}
\end{table*}

\begin{figure*}
\begin{minipage}{138mm}
\begin{center}
\subfigure[]{ \includegraphics[width=6.5cm ,height=6.5cm]{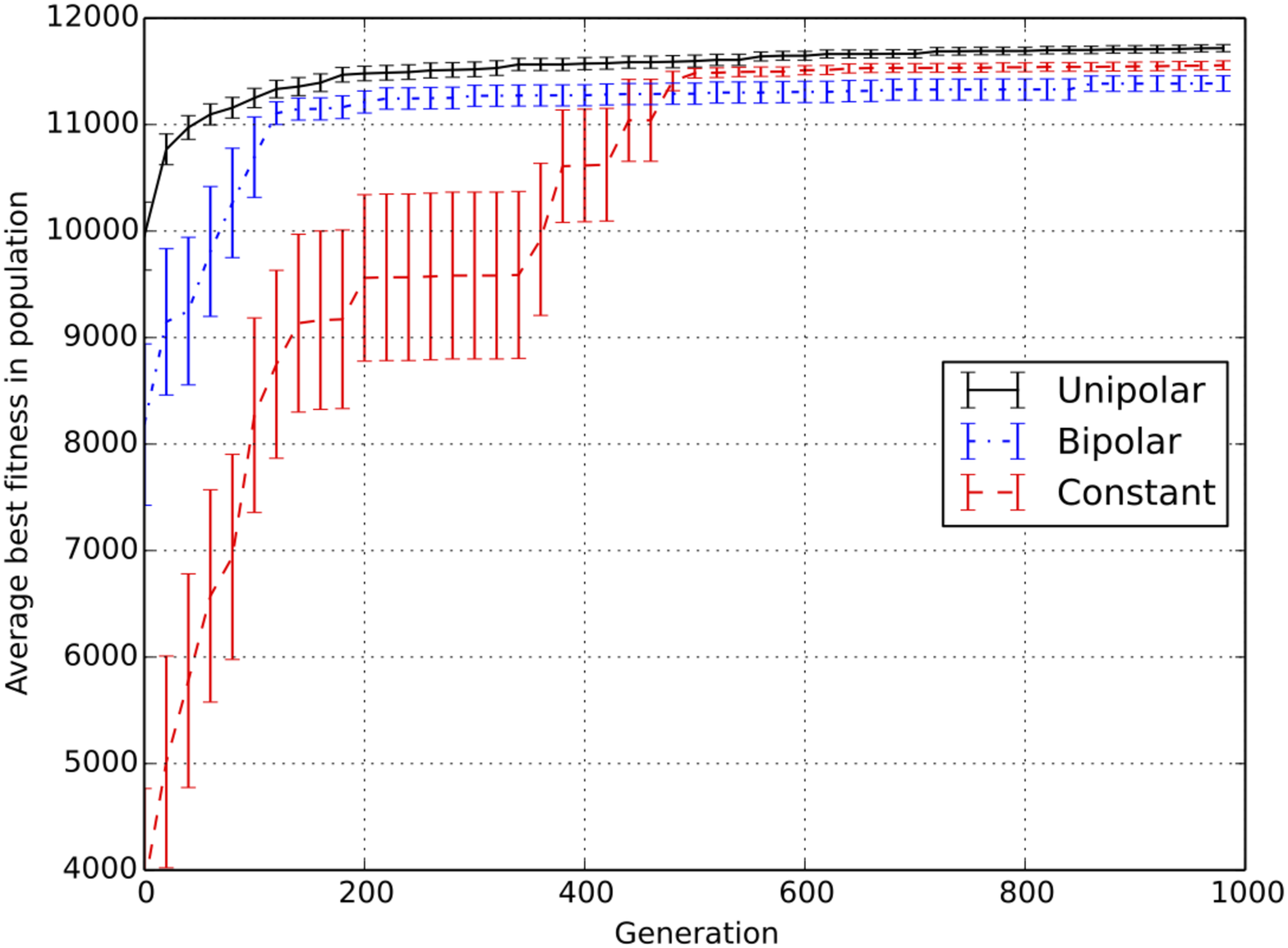}}
\subfigure[]{ \includegraphics[width=6.5cm ,height=6.5cm]{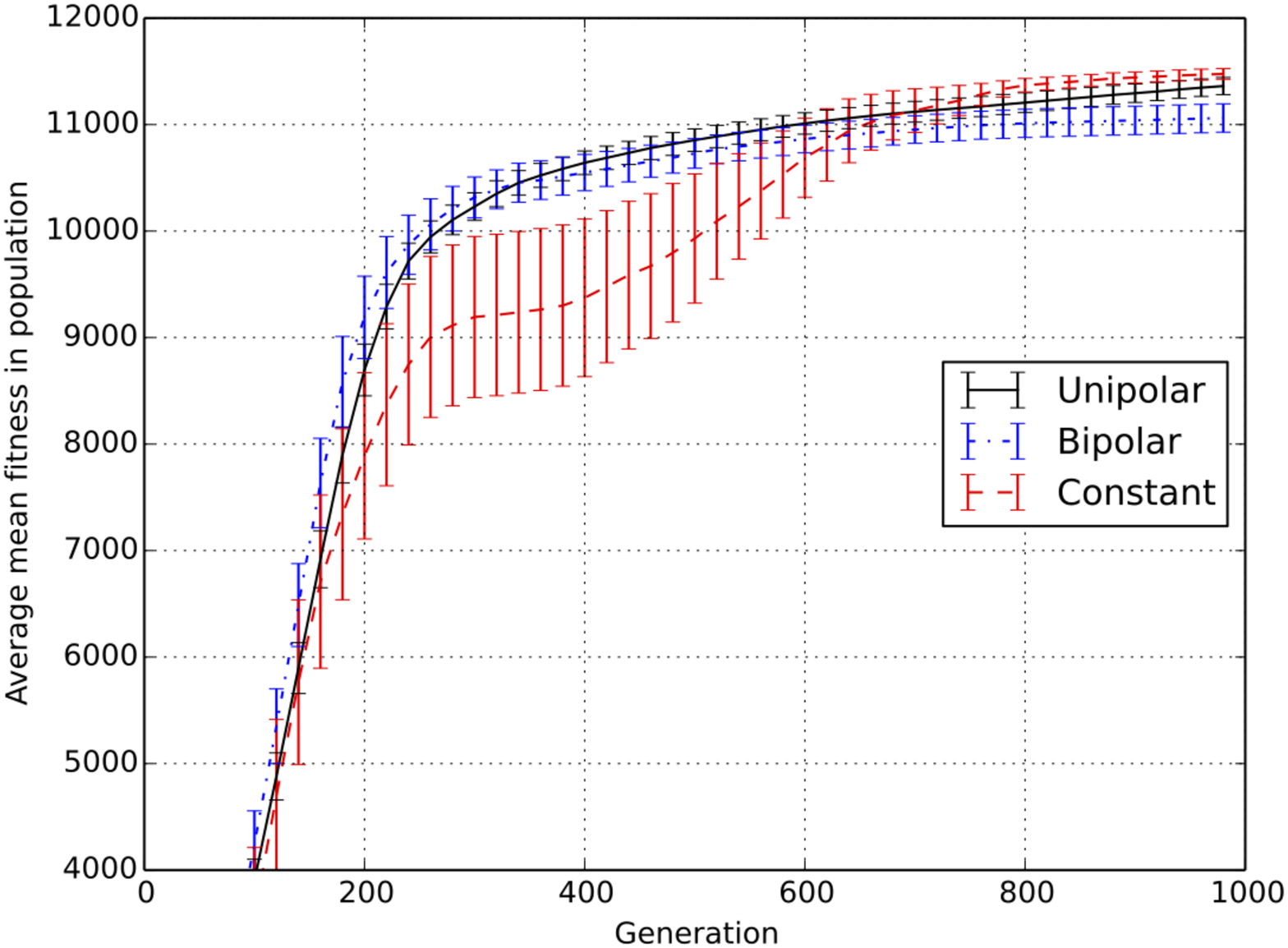}}\\
\subfigure[]{ \includegraphics[width=6.5cm ,height=6.5cm]{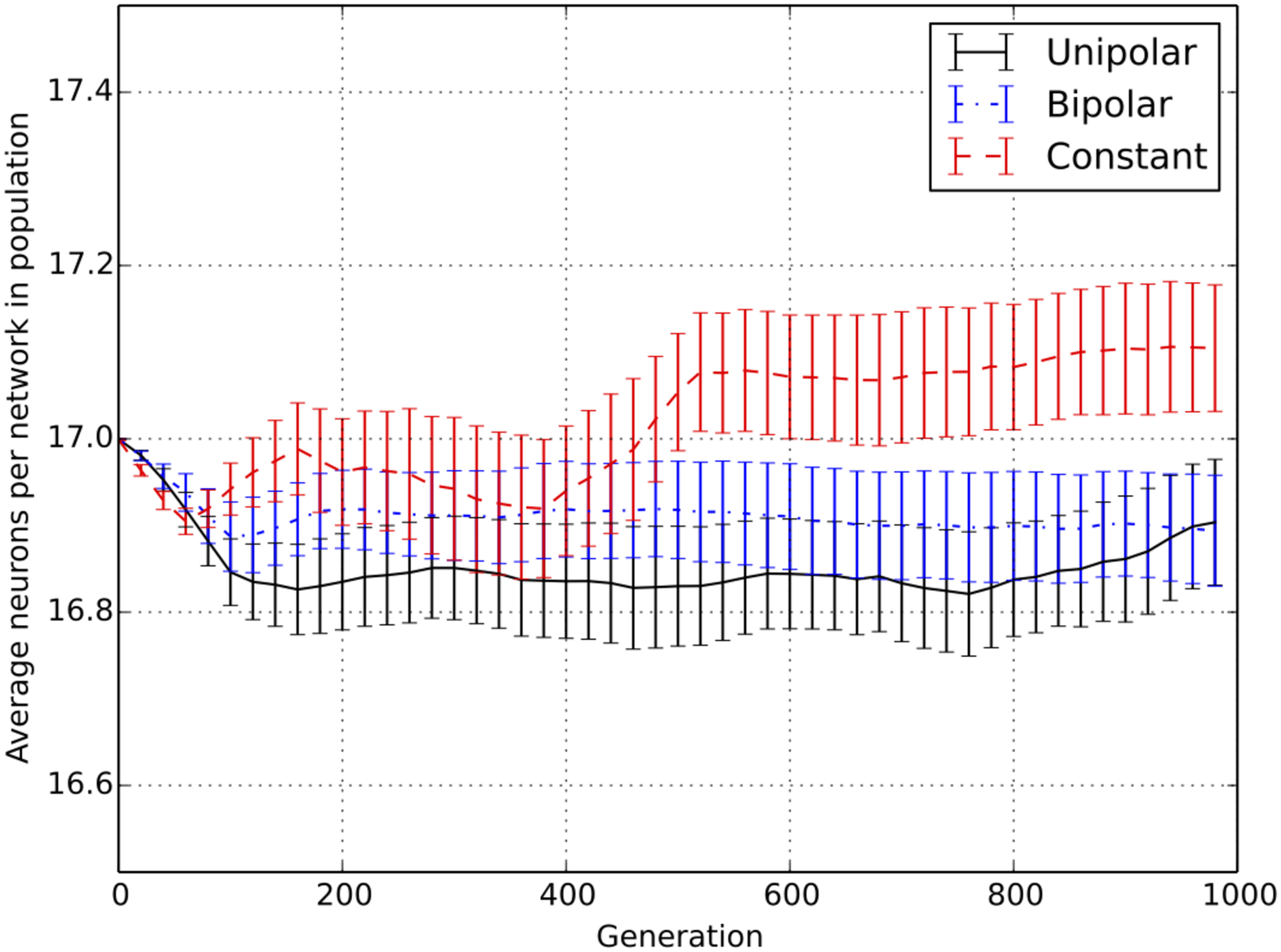}}
\subfigure[]{ \includegraphics[width=6.5cm ,height=6.5cm]{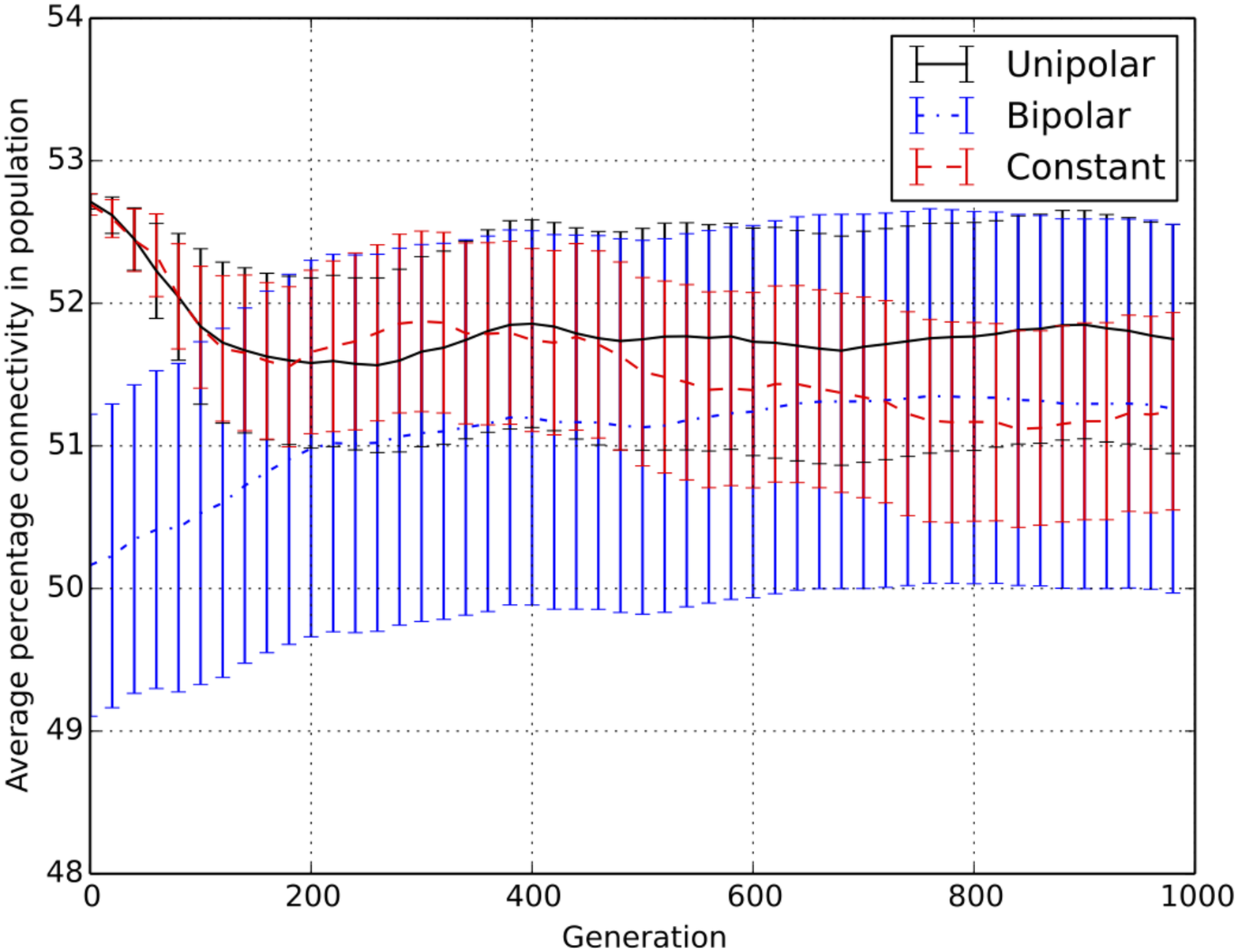}}
\end{center}
\caption[]{\label{figure5}Phototaxis mean (a) best fitness (b) avg. fitness (c) hidden layer nodes (d) percentage connectivity for unipolar, bipolar, and constant synapse networks. Bars denote standard error.}
\end{minipage}
\end{figure*}

Self-adaptive mutation parameters (Figure~\ref{figure6}, Table~\ref{table2}) do not statistically differ between the synapse types, but are all statistically different for different values within a synapse type.  This shows some context-sensitivity as the parameters automatically find suitable values to allow for the successful evolution of succesful networks.  In particular, Figure~\ref{figure6}(c) shows a smoother profile and faster convergence for $\tau$ (the rate of connection selection) for unipolar and bipolar networks compared to constant networks.  We note that the higher rate for constant networks, and specifically the increase in $\tau$ from generations $\approx$400-500 permits more search and corresponds to the jump in fitness evidenced in Figure~\ref{figure5}(a).  If nothing else, this demonstrates a very ``direct'' example of how self-adaptive search rates are used to drive network exploration to find fitter solutions.

\begin{table*}
\caption{Phototaxis averages and standard deviations for mutation parameters for the three synapse types. Symbols indicate the value is statistically (p$<$0.05) higher than $^{o}$ = Unipolar $\dagger$ = Bipolar, * = Constant.}
\begin{tabular}{@{}lcccc}
     			 & $\mu$       	&  $\psi$     		& $\omega$   	& $\tau$        \\ 
Unipolar		&NA				& 0.065 (0.03)		&0.092 (0.03)	&0.01 (0.01)	\\ 
Bipolar			&NA				& 0.052 (0.02)		& 0.135 (0.07)	&0.01 (0.01) 	\\	
Constant		&0.018 (0.01)	& 0.056 (0.03)		&0.122 (0.09)	&0.011  (0.01)	\\ 
\end{tabular}
\label{table2} 
\end{table*}

\begin{figure*}
\begin{minipage}{138mm}
\begin{center}
\subfigure[]{ \includegraphics[width=4cm ,height=4cm]{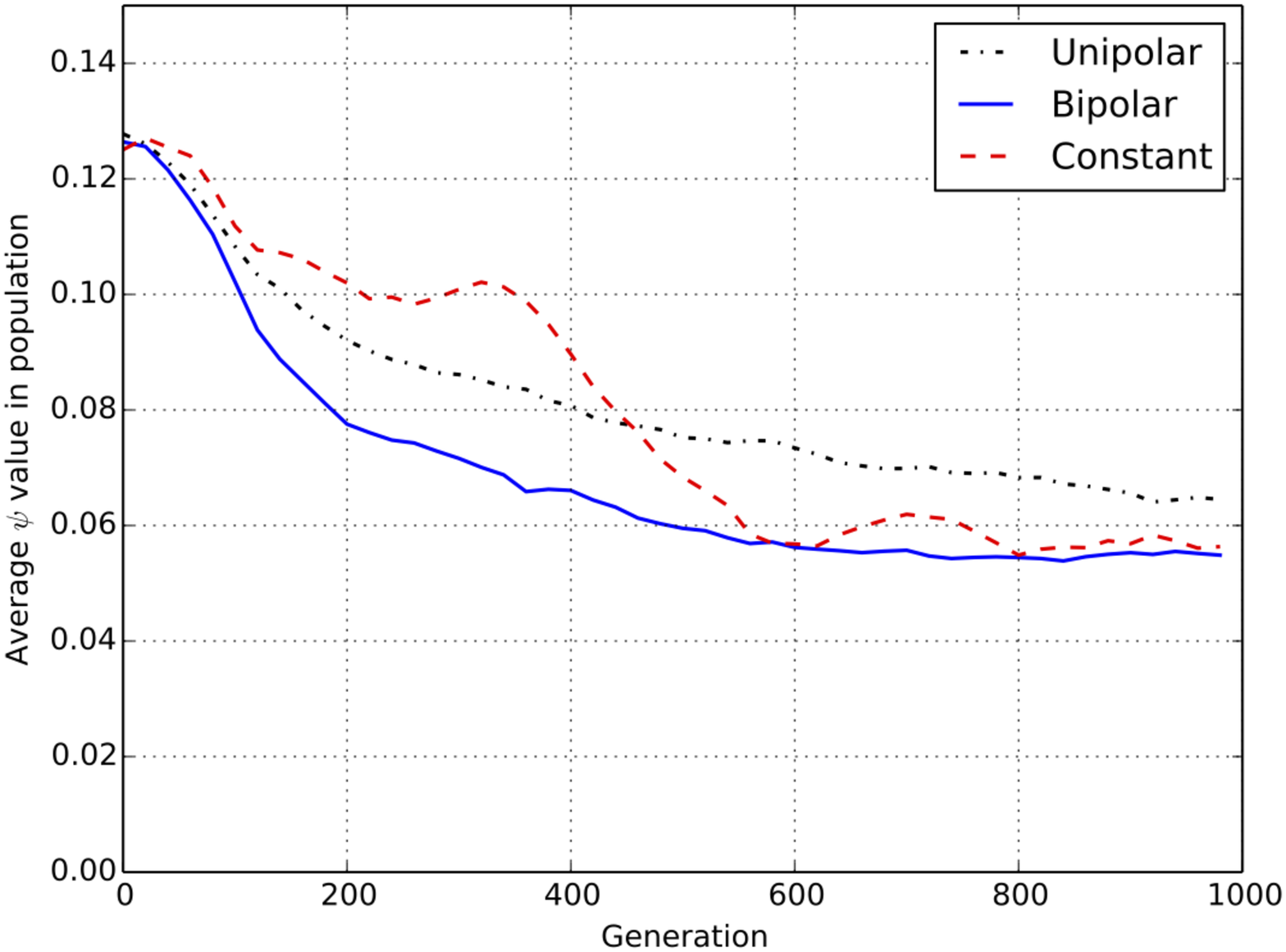}}
\subfigure[]{ \includegraphics[width=4cm ,height=4cm]{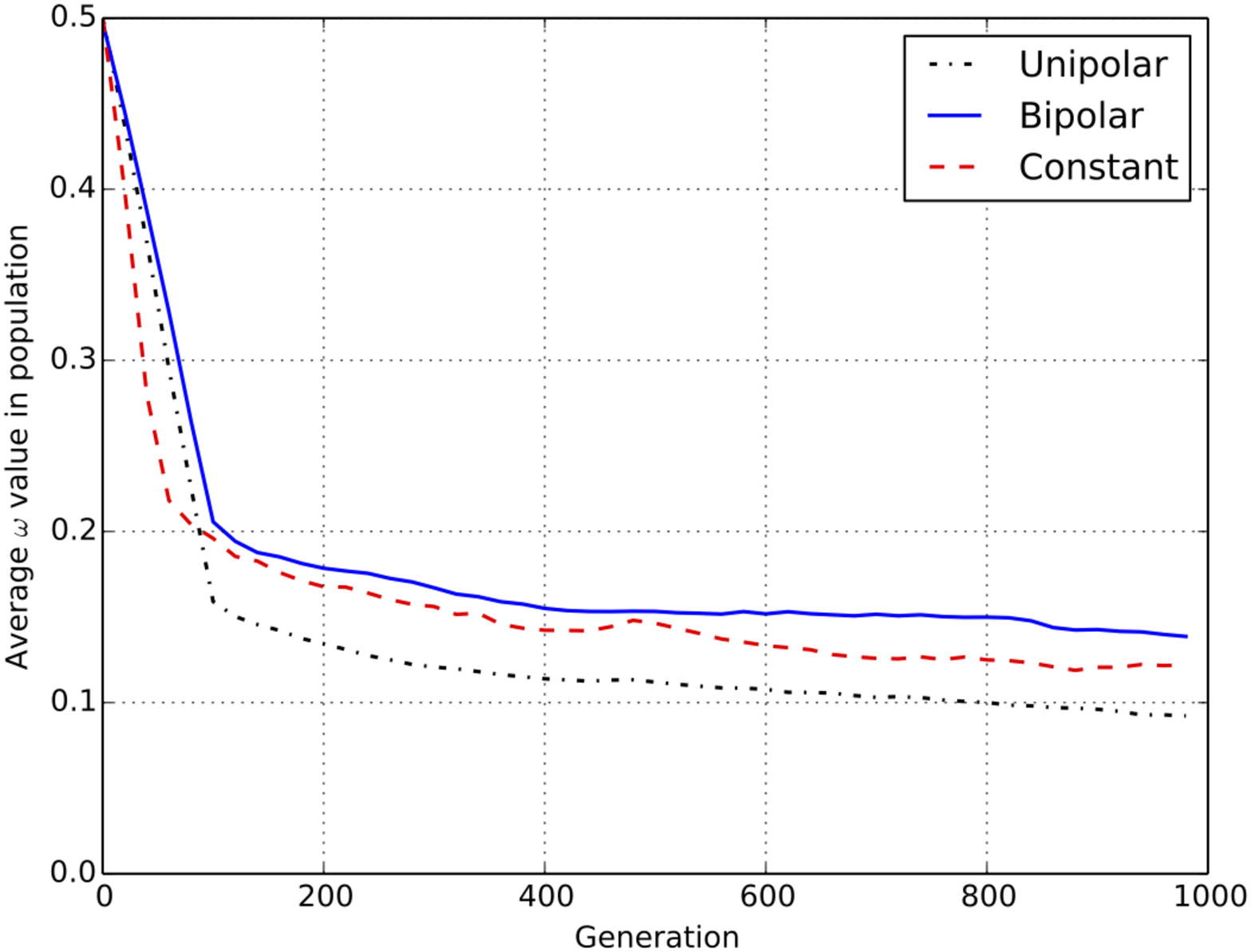}}
\subfigure[]{ \includegraphics[width=4cm ,height=4cm]{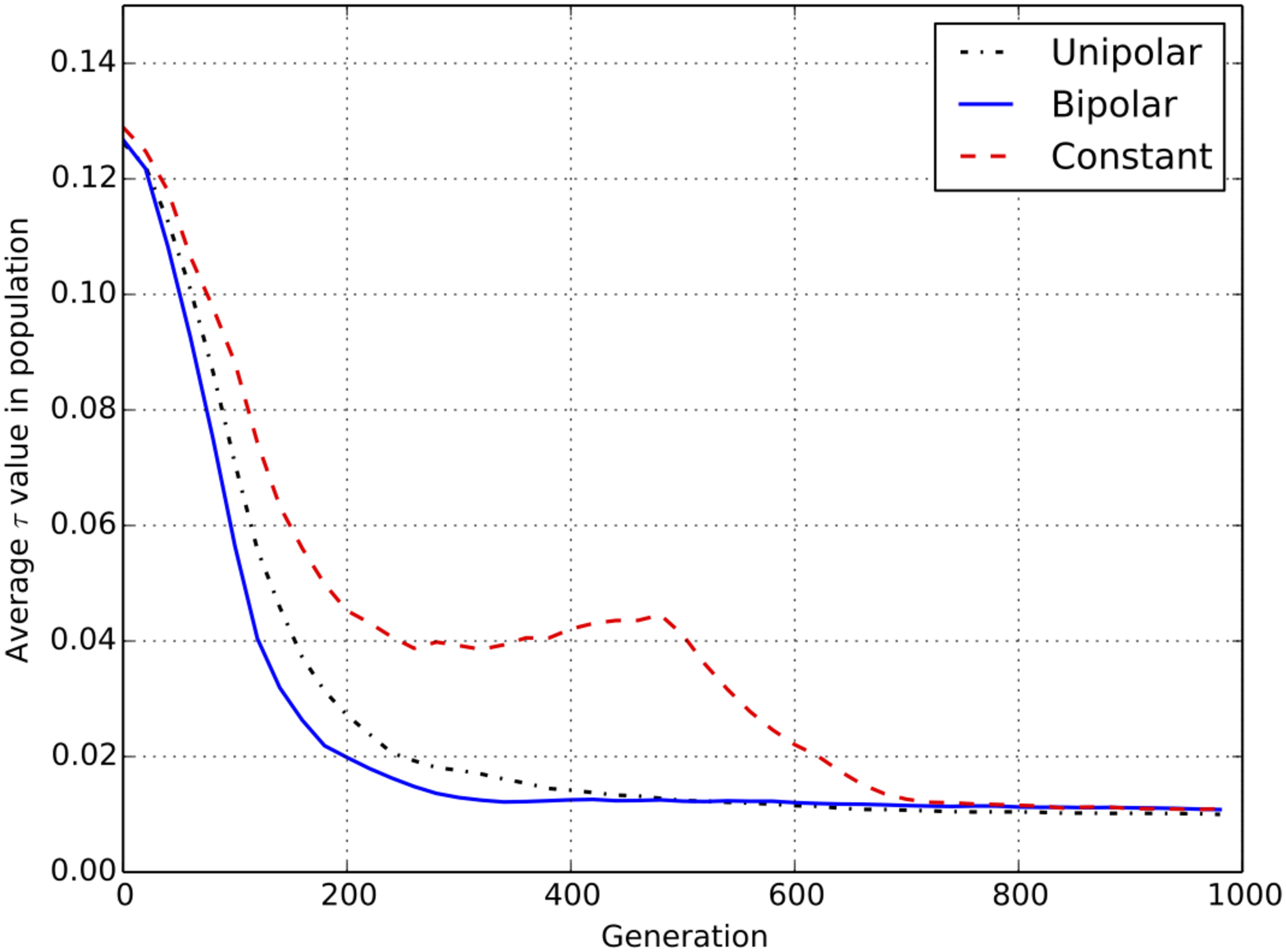}}
\end{center}
\caption[]{\label{figure6}Phototaxis mean (a) node addition/removal event rate $\psi$ (b) node addition rate $\omega$ (c) connection addition/removal rate $\tau$ for unipolar, bipolar, and constant synapse networks.}
\end{minipage}
\end{figure*}

\subsection{Results II: Tmaze}

Table~\ref{table3} shows that the unipolar memristor networks are able to find solutions to the test problems in significantly fewer generations than both the bipolar memristor and constant connection networks.  The unipolar memristors can only be in two states, and as such the possible network attractor space is significantly more constrained during a trial than that of the bipolar network.  As the GA is responsible for setting up useful network activity --- attractors that produce pathfinding behaviour --- the relationship between topology and in-trial behaviour can be more expediently explored.  The plasticity provided by the unipolar network is still useful for behaviour generation, hence the significant speedup over constant connection networks.  The role of plasticity will be explored further in Section~\ref{plasticity}.

Best and average fitness values (Table ~\ref{table3}) do not vary significantly between the network types.  This indicates that the unipolar memristor can generate competitive pathfinding behaviour in addition to adapting significantly faster to the dynamic T-maze. Figure~\ref{figure7}(a) shows the best fitness for constant networks always lagging behind those of the other network types, and shows that the final order is the same as the order of synapse complexity.  The more homogeneous starting fitness values (compare Figure~\ref{figure5}(a)) are due to the allocation of fitness in the networks, which receive a value of 8000 if neither reward zone is reached, showing that none of the network types contain even partial solutions in their initial populations.  Average fitness (Figure~\ref{figure7}(b)) shows that unipolar networks gain a fitness advantage in the first $\approx$200 generations which is ceded in the final 200 generations.

\begin{table*}
\caption{ T-maze averages and standard deviations for controller parameters for the three synapse types. Symbols indicate the value is statistically (p$<$0.05) better than $^{o}$ = Unipolar $\dagger$ = Bipolar, * = Constant.}
 \begin{tabular}{@{}lcccccc}
      		 & Best fit       &  Avg fit       & Gens. to solve    & Nodes        &Connectivity     \\ 
    Unipolar & 1602.5 (422)   & 3115.0 (631)   & 12.47 $\dagger$*(10.7)   & 17.01 (0.62) & 52.98 (3.7) \\
 	Bipolar  & 1368.3 (806)   & 2679.6 (966)   & 47.93 (70.8)   & 17.07 (0.56) & 51.7 (5.98)   \\ 
 	Constant & 1671.6 (656)   & 2837.1 (1284)  & 27.73 (38.9)   & 16.9 (0.69)  & 52.38 (4.32)  \\ 
\end{tabular}
\label{table3}
\end{table*}

\begin{figure*}
\begin{minipage}{138mm}
\begin{center}
\subfigure[]{ \includegraphics[width=6.5cm ,height=6.5cm]{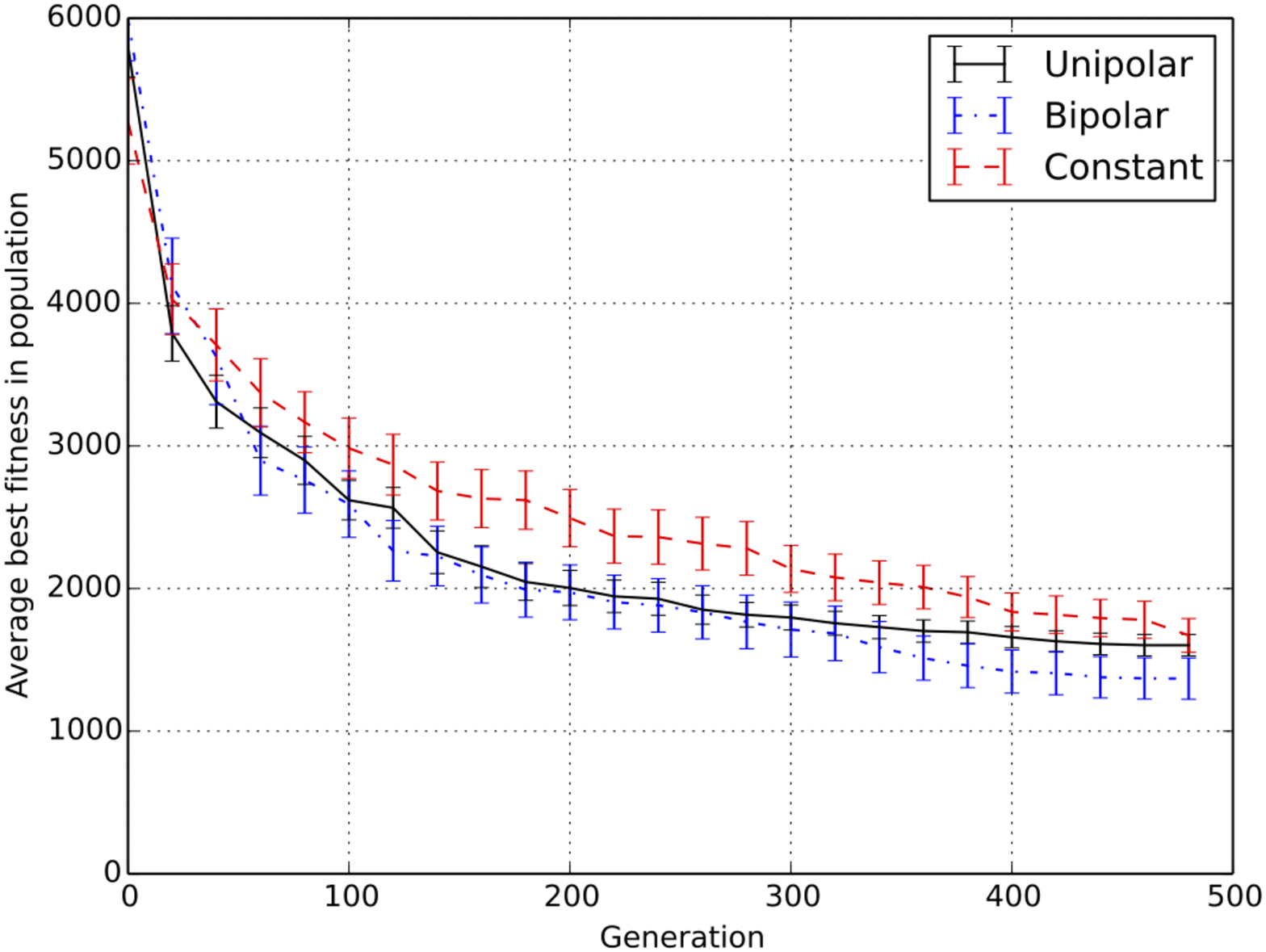}}
\subfigure[]{ \includegraphics[width=6.5cm ,height=6.5cm]{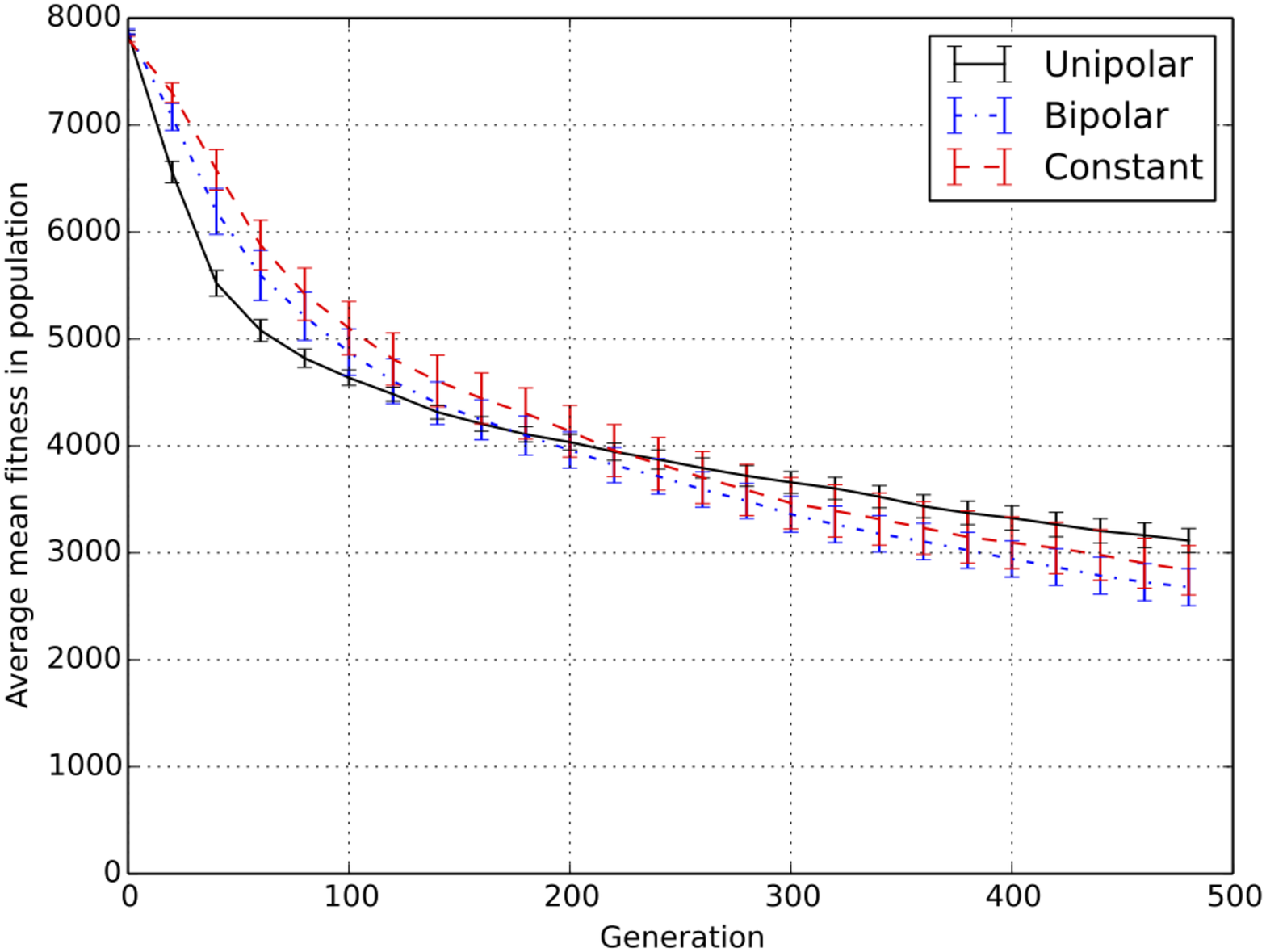}}\\
\subfigure[]{ \includegraphics[width=6.5cm ,height=6.5cm]{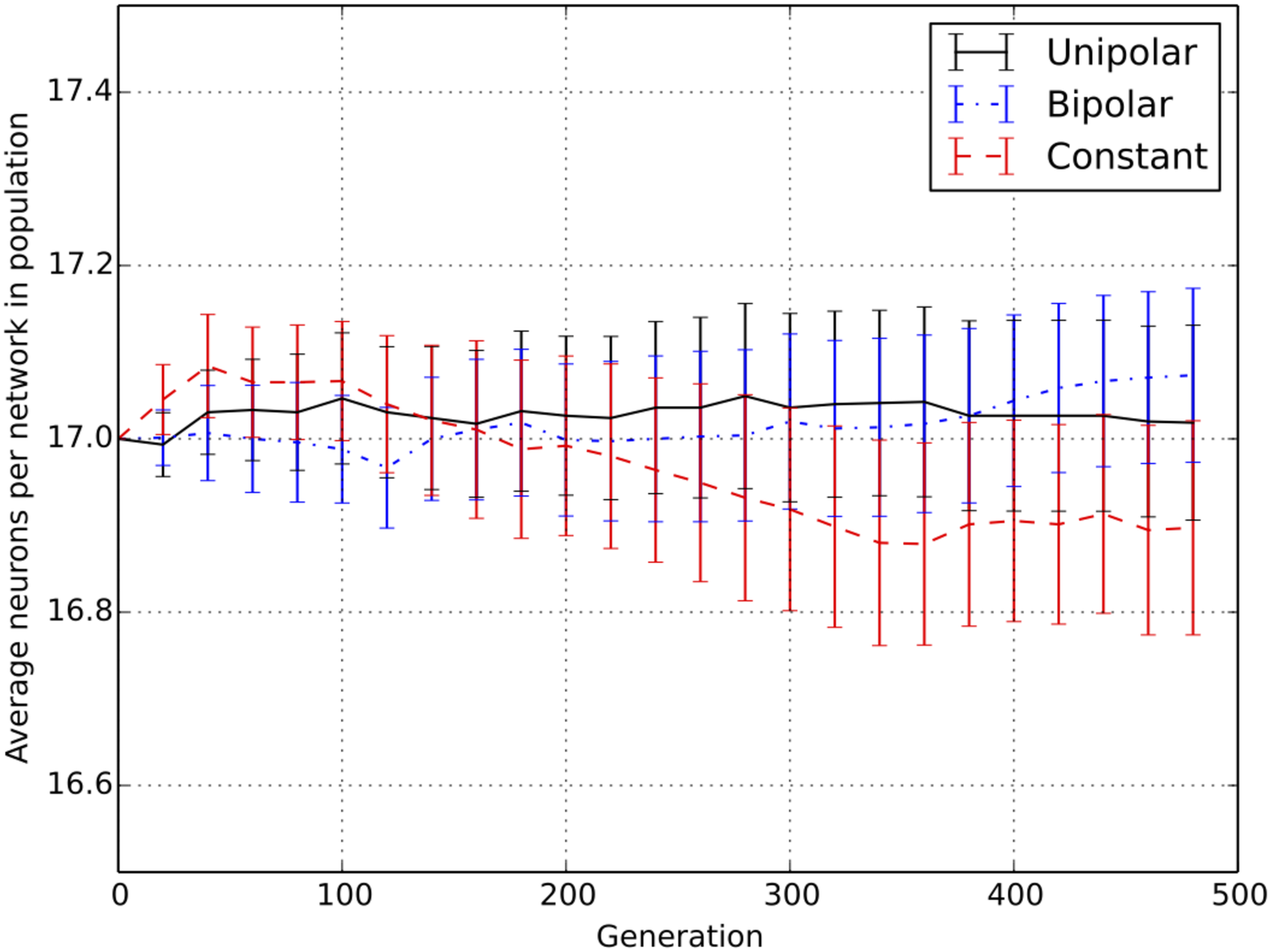}}
\subfigure[]{ \includegraphics[width=6.5cm ,height=6.5cm]{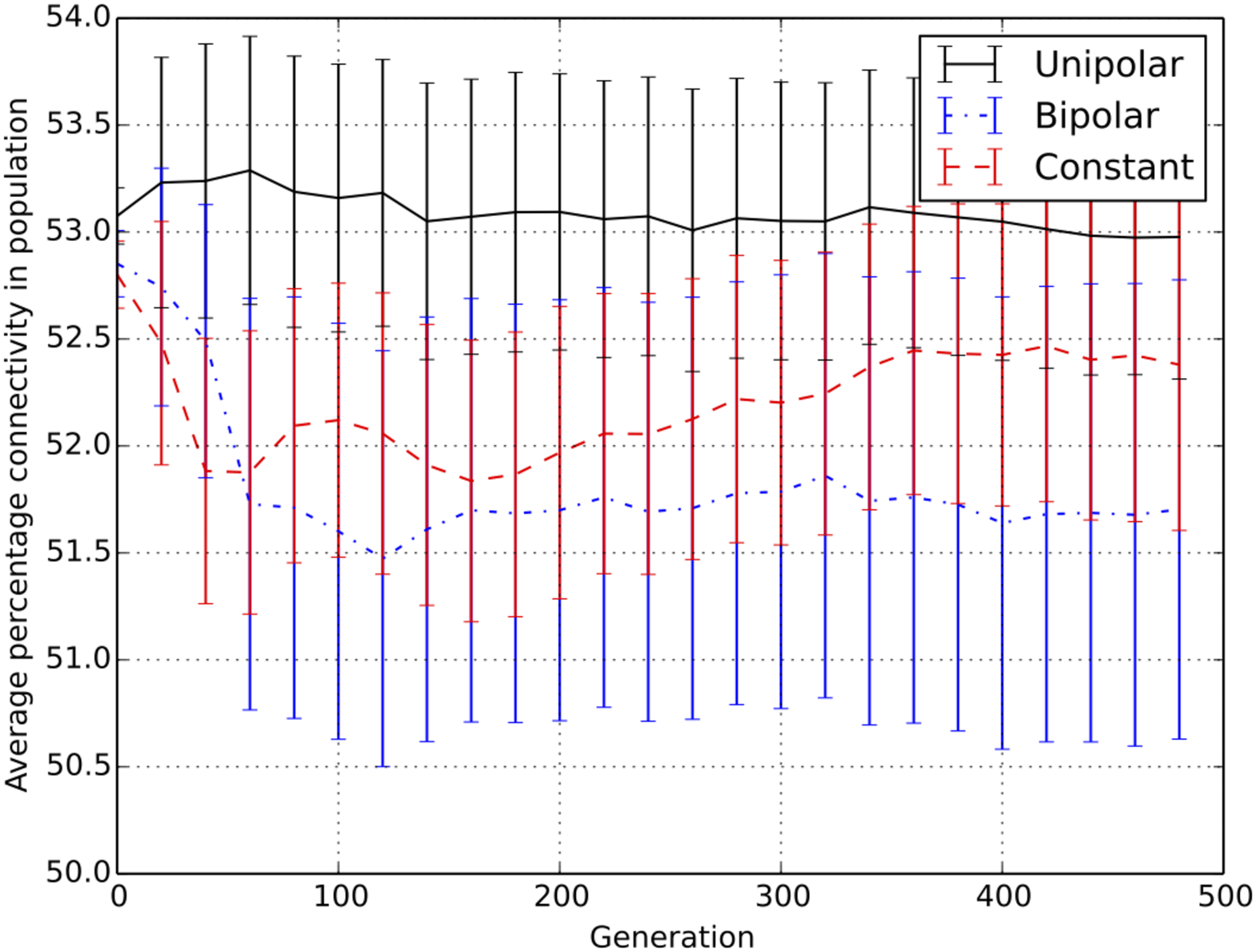}}
\end{center}
\caption[]{\label{figure7}T-maze mean (a) best fitness (b) avg. fitness (c) hidden layer nodes (d) percentage connectivity for unipolar, bipolar, and constant synapse networks. Bars denote standard error.}
\end{minipage}
\end{figure*}

In terms of topology (the last two columns of Table~\ref{table3}), the numbers of connected hidden layer nodes (Figure~\ref{figure7}(c)) and connections (Figure~\ref{figure7}(d)) do not vary significantly between the three network types.  We note that the standard error is very high for the number of enabled connections; this variance is the reason that no statictical significance is observed.  

Self-adaptive parameters (Figure~\ref{figure8}, Table~\ref{table4}) are again shown to be context-sensitive.  For the same network type, the parameters are statistically different from each other (compare Figure~\ref{figure6}, Table~\ref{table2}.  Across the networks, $\psi$ (rate of node addition/removal events) was statistically higher in unipolar networks than either bipolar or constant networks.  $\tau$ (rate of connection addition/removal) has statistically higher in unipolar networks (avg. 0.049) than in constant networks (avg. 0.024) --- Table~\ref{table4}.  Unipolar networks appear to require more genetic search of connection space in this task.\\

\begin{table*}
\caption{T-maze averages and standard deviations for mutation parameters for the three synapse types.  Symbols indicate the value is statistically (p$<$0.05) higher than $^{o}$ = Unipolar $\dagger$ = Bipolar, * = Constant.}
\begin{tabular}{@{}lcccc}
    & $\mu$       	&  $\psi$      & $\omega$   		& $\tau$         \\ 
    Unipolar & NA   		& 0.11 $\dagger$*(0.03)   & 0.358 (0.18)   	& 0.049 *(0.04)  \\
 	Bipolar  & NA   		& 0.09 (0.03)   & 0.315 (0.17)   	& 0.04 (0.03)    \\
 	Constant & 0.044 (0.03) & 0.074 (0.04)  & 0.33 (0.22)   	& 0.033 (0.02)   	\\ 
\end{tabular}
\label{table4}
\end{table*}

\begin{figure*}
\begin{minipage}{138mm}
\begin{center}
\subfigure[]{ \includegraphics[width=4cm ,height=4cm]{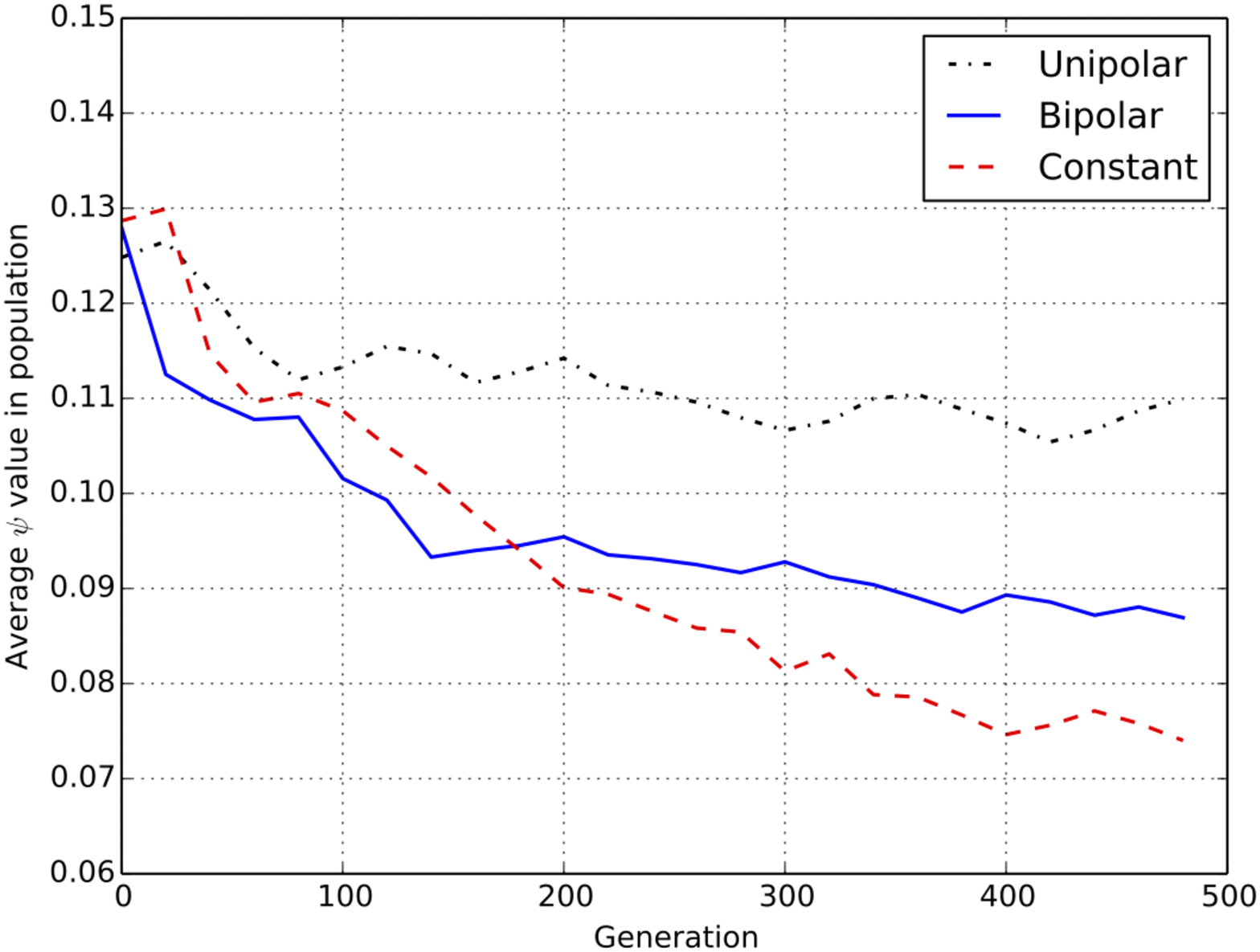}}
\subfigure[]{ \includegraphics[width=4cm ,height=4cm]{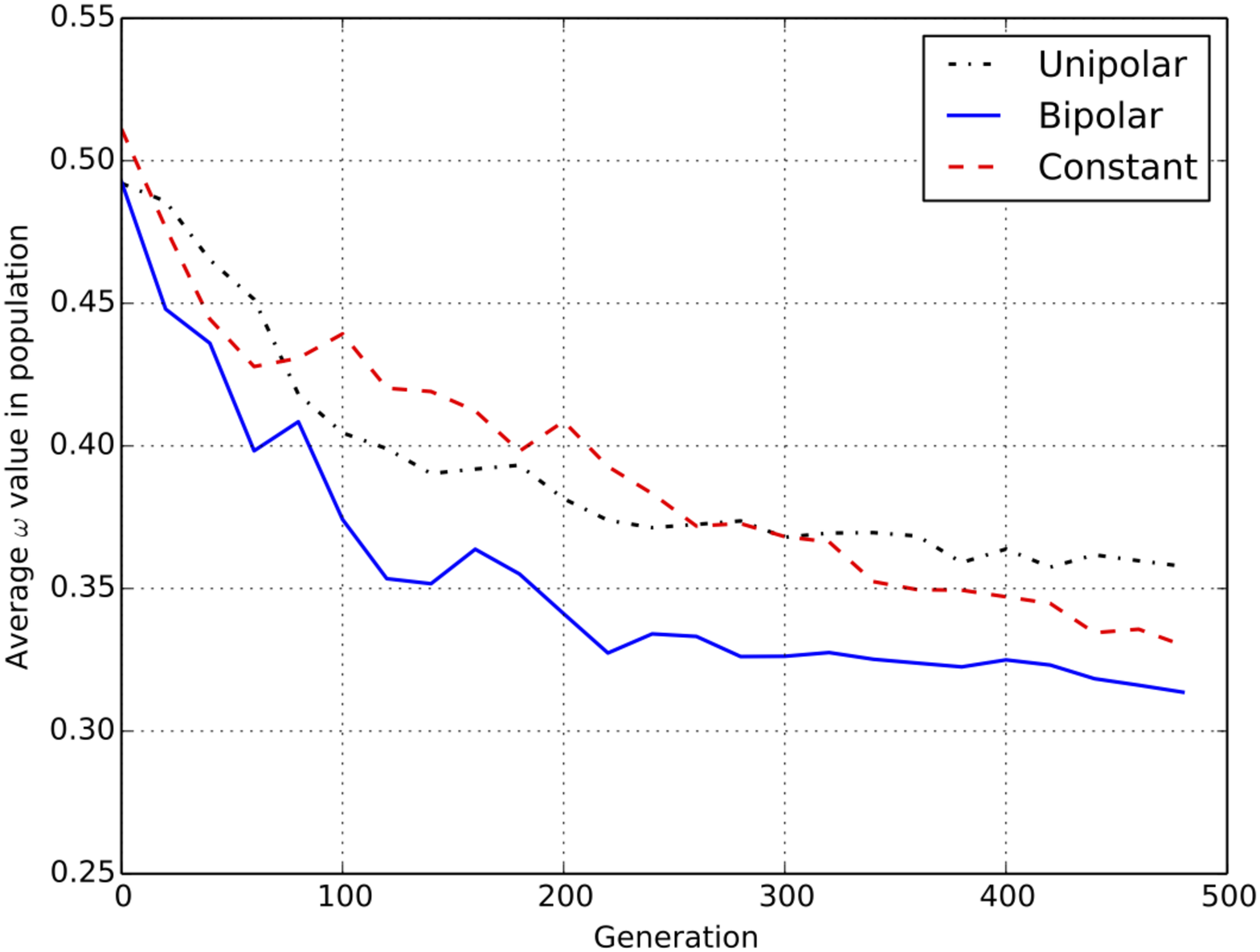}}
\subfigure[]{ \includegraphics[width=4cm ,height=4cm]{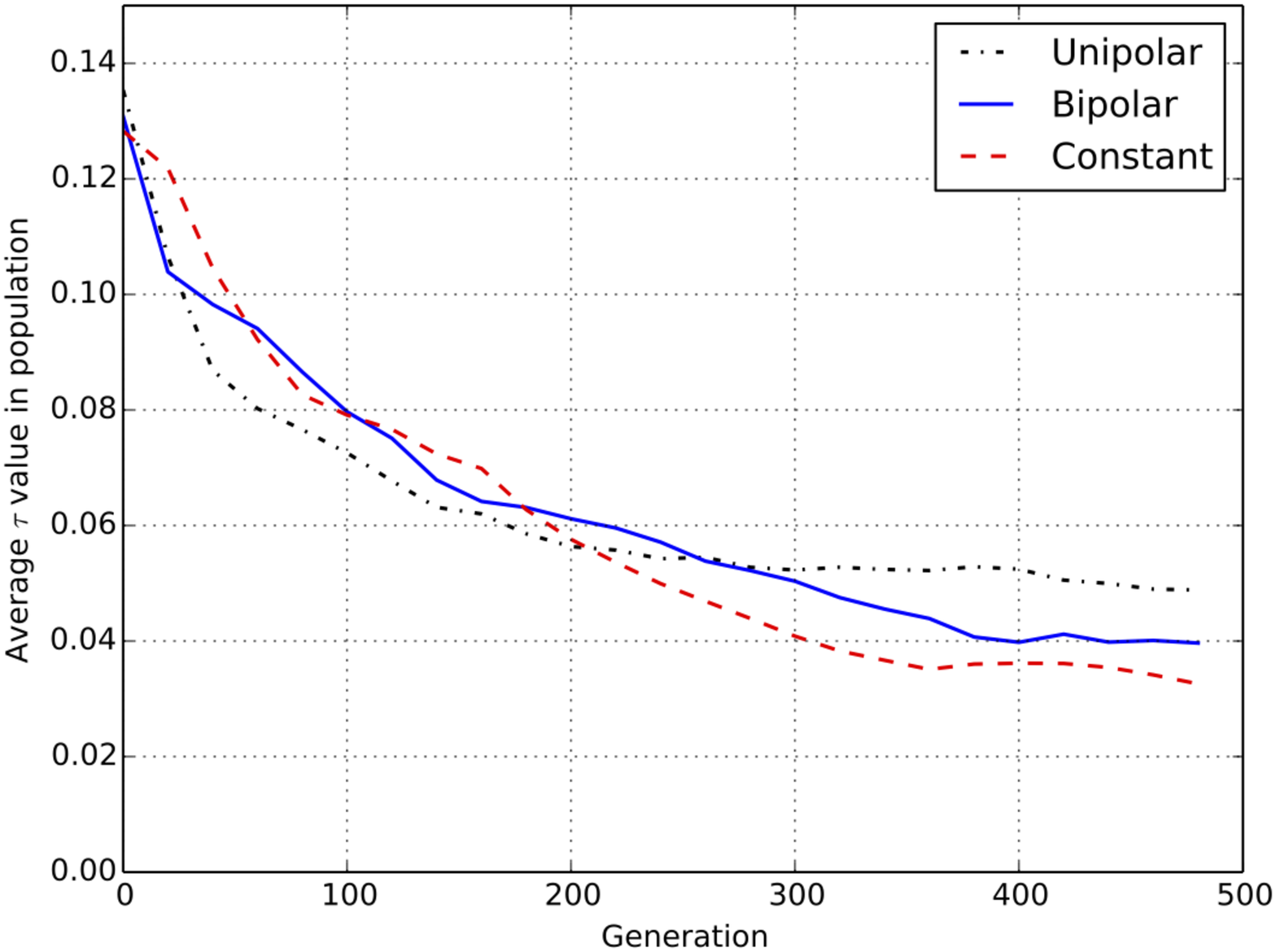}}
\end{center}
\caption[]{\label{figure8}T-maze mean (a) node addition/removal event rate $\psi$ (b) node addition rate $\omega$ (c) connection addition/removal rate $\tau$ for unipolar, bipolar, and constant synapse networks.}
\end{minipage}
\end{figure*}

\subsection{Synaptic Plasticity}
\label{plasticity}

Plasticity was seen to effect both key indicators of controller performance: fitness and the number of generations taken to ``solve'' the task.  For fitness,  plasticity is used as a way of flexibly generating the required action from an arbitrary sequence of input states.  Action sequences are seen to be more heterogeneous in the unipolar case as a single switch can cause a large peturbation in network activity, with constant connections being the most homogeneous in this regard.  For the number of generations to solve, plasticity allows some degree of behavioural exploration to take place online, removing the onus from the GA and increasing adaptivity.

Unipolar networks used the rapid-switching ability of the synapse in two main ways: (i) to perform online ``connection selection'' e.g. to switch a synapse to a given state once and leave it there, and (ii) varying the connectivity map of the network multiple times during a trial to create weight oscillators in the network, whereby the firing on the neurons and switching of the synapses synchronises through time to generate appropriate output actions from a subgroup of neurons.

It was initially thought that binary nature of unipolar resistance switching would lead to ``twitchy'' controller behaviour.  Some of the less fit/early generation unipolar networks showed noticable oscillations in path generation, but later, fitter networks were shown to avoid this problem by synchronising switching between two synapses to the same neuron, e.g. where one is in the LRS and the other is in the HRS simulataneously, the receiving neuron receives a constant input of 0.1 + 0.9 / 2 = 0.5.  By switching at the same time, this constant input can be preserved, and used to stabilise the network making behaviour generation easier.  In other words, the quality of generated paths is not significantly impeded by the simpler binary switching nature of the synapse when compared to bipolar memristors.\\

\begin{figure*}
\begin{minipage}{138mm}
\begin{center}
\subfigure[]{ \includegraphics[width=4cm ,height=4cm]{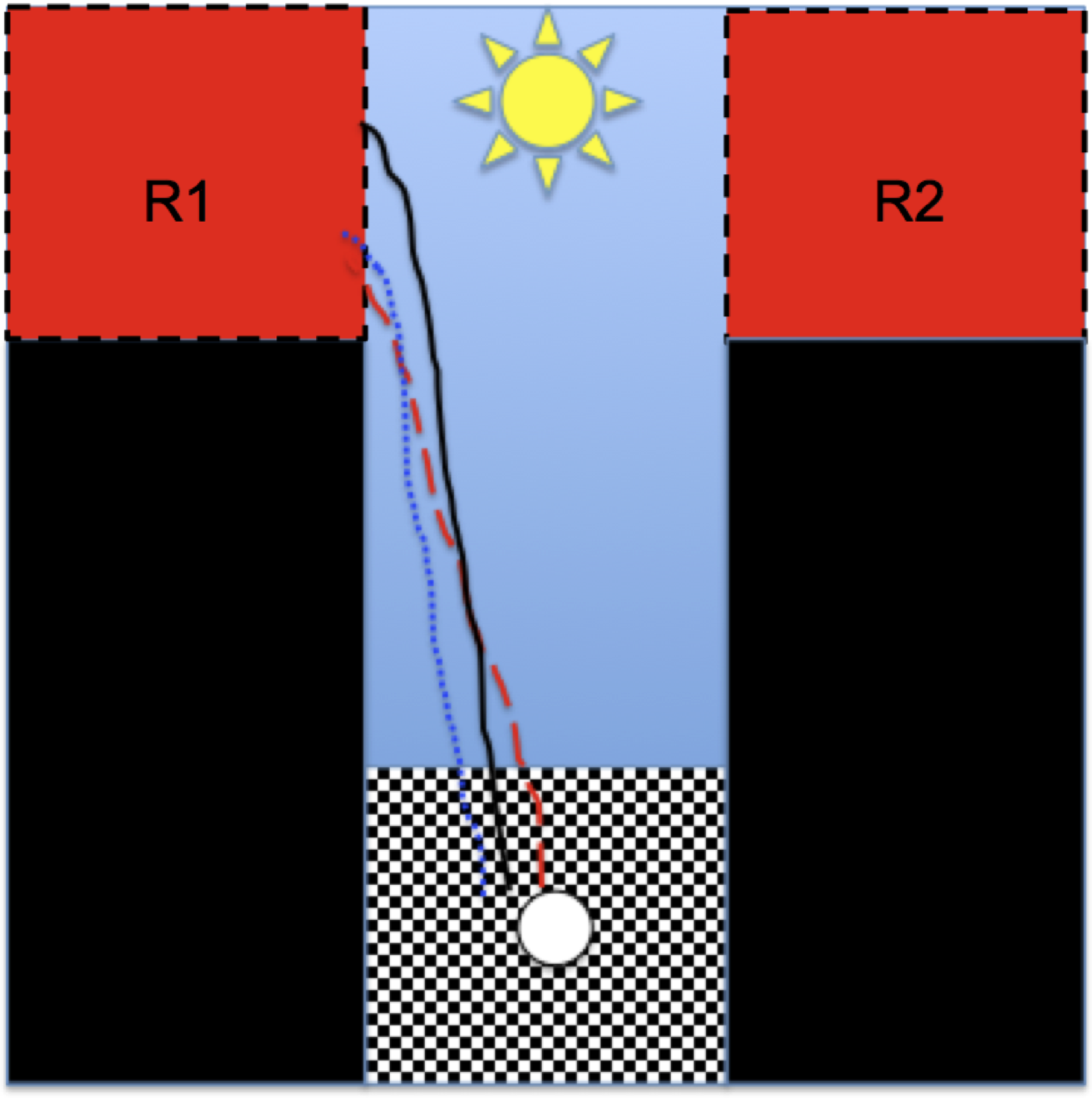}}
\subfigure[]{ \includegraphics[width=4cm ,height=4cm]{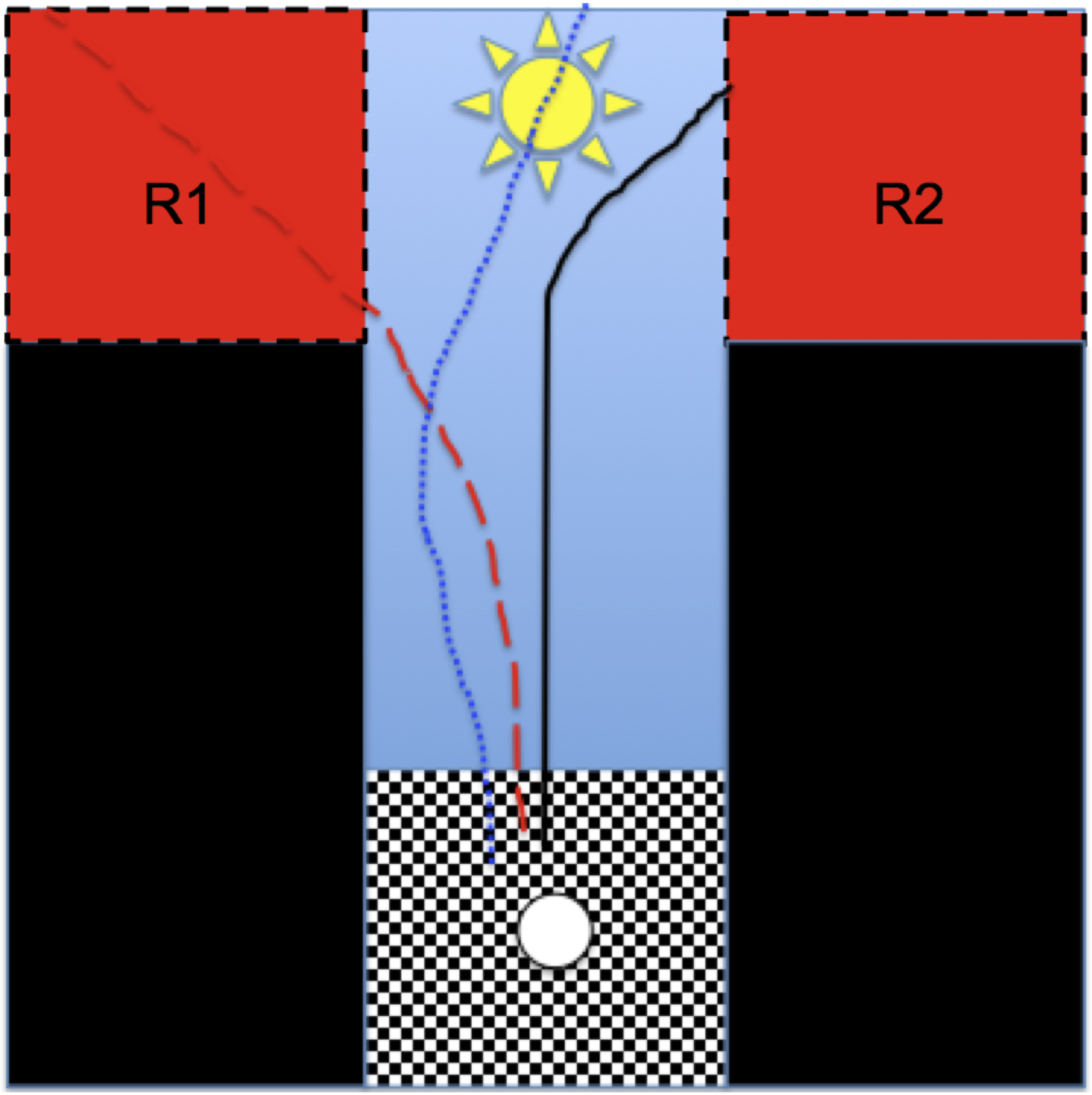}}
\subfigure[]{ \includegraphics[width=4cm ,height=4cm]{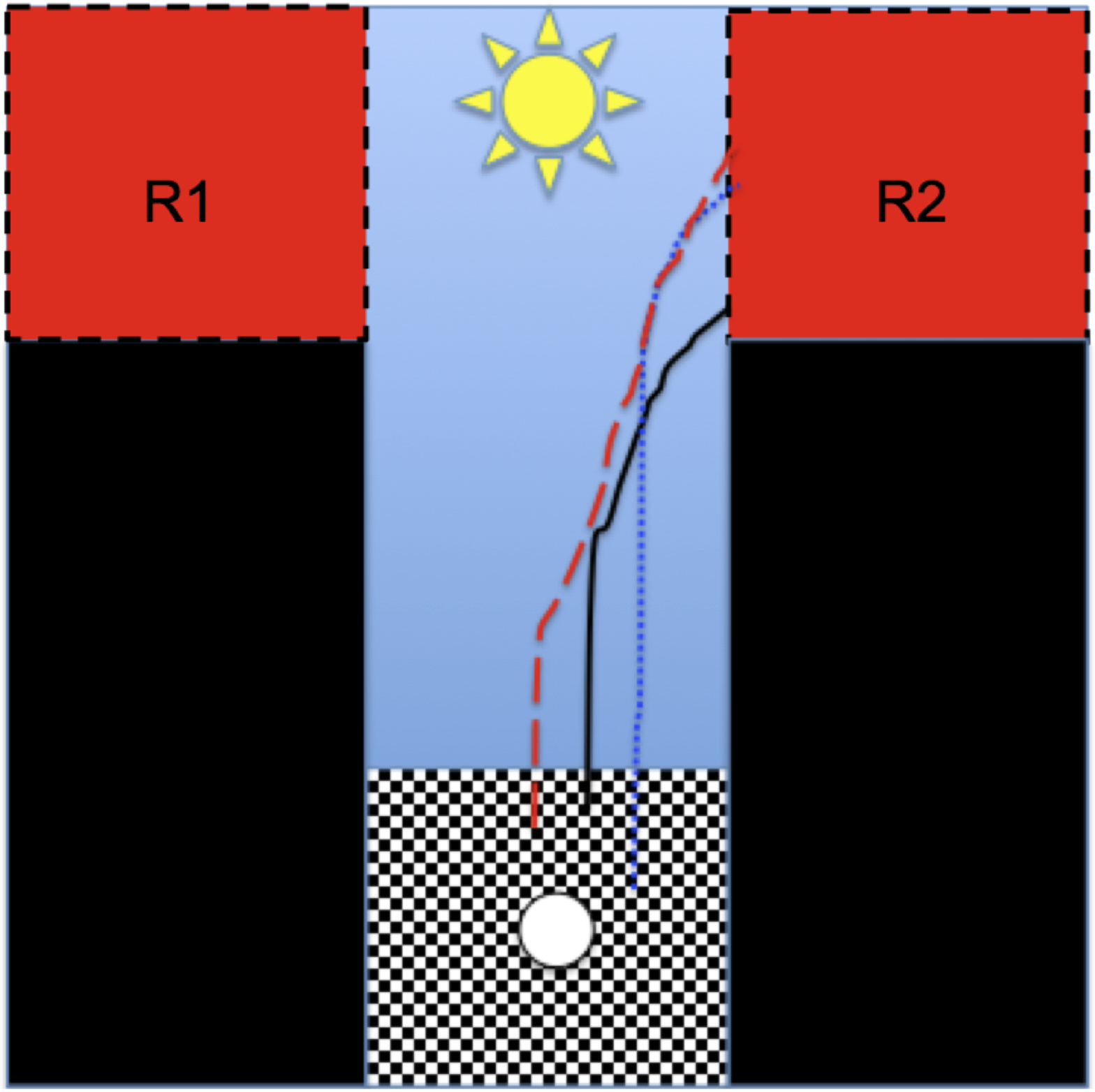}}
\end{center}
\caption[]{\label{figure9} Showing adaptation in the T-maze environment.  For the best controller of each synapse type, (a) shows the final path for that controller in finding R1 before the reward zone is switched to R2.  Note that all paths are approximately equally good.  (b) 15 generations after the reward zone switch, the best unipolar network uses online plasticity adapted to the new reward zone.  Binary switching allows a diverse range of behaviours to be explored online, reducing the number of generations required to adapt.  The bipolar memristor similarly uses online adaptation to search a behavioural repertoire, but the gradual analog plasticity variation means that the potential behaviours are less diverse than in the unipolar case.  The constant connection can only adapt through GA application, and as such it's best path is similar to the final path for finding R1.  The best controllers in this case are those that come closest to finding R2. (c) After 50 generations, all controllers have adapted to the new environment.   Paths are: unipolar (black solid line), bipolar (blue dotted line), and constant (red dashed line).}
\end{minipage}
\end{figure*}

Figure~\ref{figure9} shows how plasticity gives rise to fast adaptation in the T-maze environment.  When comparing the best controller for each synapse type, it is shown that unipolar plasticity allows for a more expedient search of a larger immediate behaviour space, leading to solving the problem more quickly.  On the other hand, the constant networks have to perform all of their behaviour exploration via GA operation, and so react more slowly.  Both constant and bipolar networks initially struggle to locate the second reward zone as their behavioural exploration is more constrained than the unipolar networks.\\

\begin{figure*}
\begin{minipage}{138mm}
\begin{center}
\subfigure[]{ \includegraphics[width=6.5cm ,height=6cm]{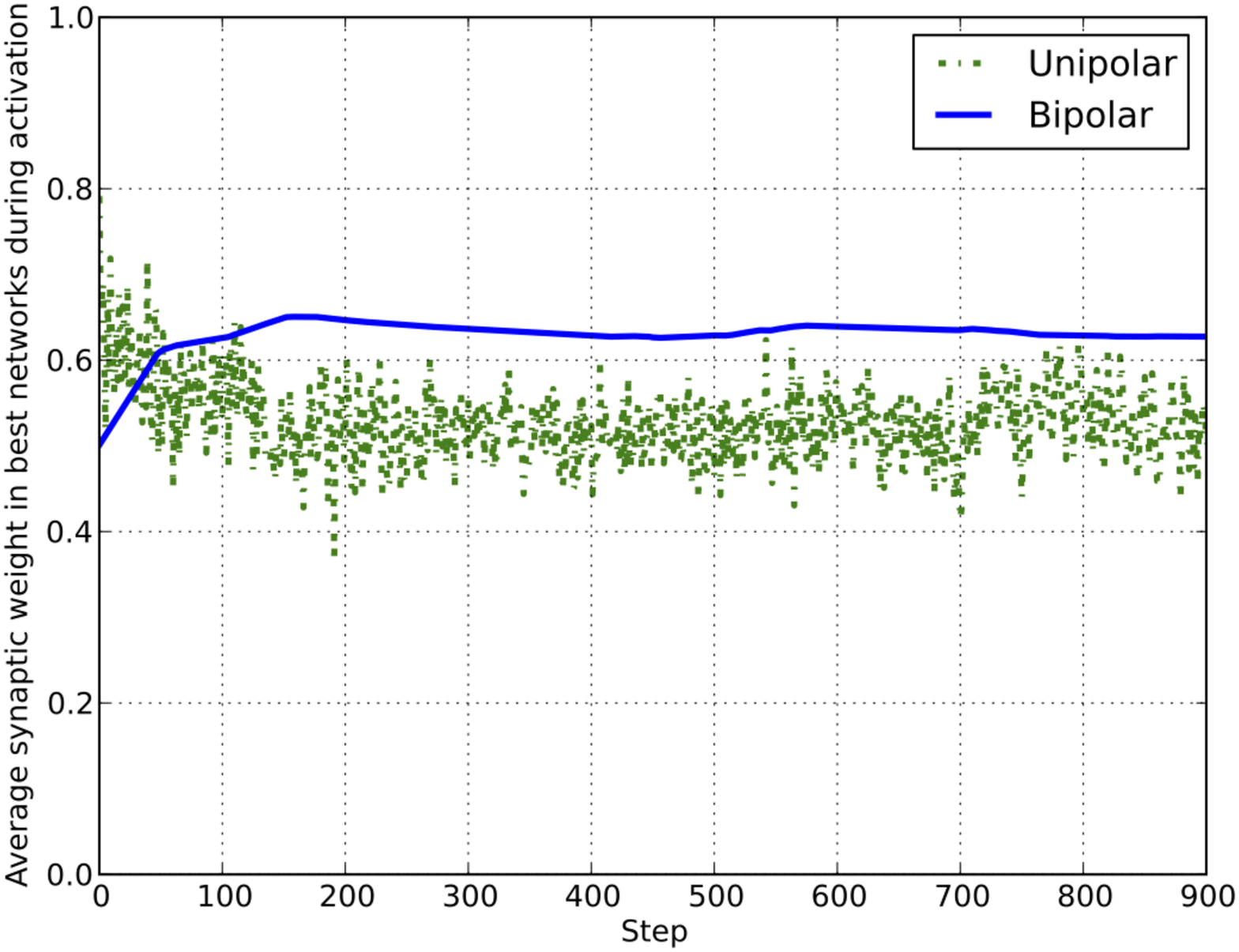}}
\subfigure[]{ \includegraphics[width=6.5cm ,height=6.5cm]{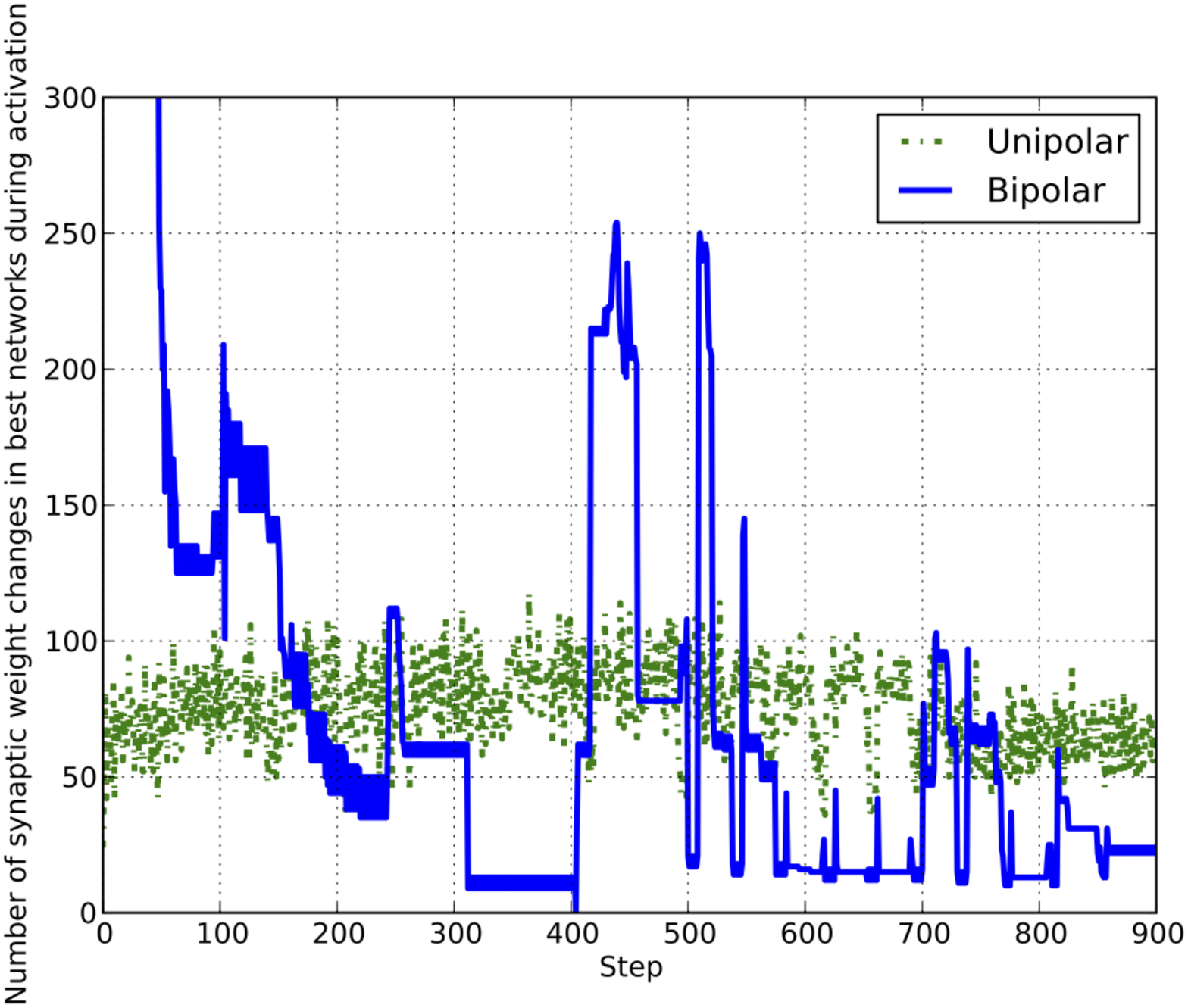}}
\caption[]{\label{figure10} For the best network of each memristor type in the T-maze, (a) average synapse weight and (b) average number of total synaptic weight changes, during network activation.  Such results are typical of each memristor type and similar between experiments.}
\end{center}
\end{minipage}
\end{figure*}

Figure~\ref{figure10}(a) shows the average synaptic weight in the best network of each memristor type during the first 900 robot steps of activation in the T-maze.  It is immediately obvious that the networks internally function differently --- average weight is much smoother in the bipolar case (avg. weight 0.528), whereas the unipolar synapses are much more distcontinuous (avg. weight 0.629).  In contrast, the total number of synaptic weight changes (Figure~\ref{figure10}(b)) over the same time period (67968 total for unipolar, 75478 total for bipolar) display a smoother profile in the unipolar case.  Fewer total switches in the unipolar case indicates that physical unipolar networks will be more long-lasting as the chemical mechanism is less likely to wear down or break (given that a unipolar switch is no more damaging to the device than a bipolar switch).   Robot steps 400-525 correspond approximately to the time where the network changes from finding R1 to finding R2.  

In the bipolar case, the increased activity in this period shows that the network is reconfiguring to produce the required behaviour.  Network stability is achieved through either spikes being sent by two neurons at the same processing step (meaning no synaptic change occurs), saturation of a synapse to a minimum/maximum (after which it is stable), or repeated pairs of positive/negative coincidence events.  To change actions, typically, a ``trigger'' neuron drives a certain {\em hidden-to-output} synapse to increase its weight in response to a change in input neuron firing (frequently) or internal network state (rarely), increasing the efficacy of spikes sent along that synapse and thus the firing frequency of the postsynaptic output neuron). Many coincidence events are required due to the more gradual nature of the weight change.  We note that such changes are more stable in the bipolar case, due to the limited effect that a synapse has on network activity in a short period of time.  Similar results are reported by~\cite{howardTEC}.

The unipolar networks are based on the concept of setting up ``weight oscillators'' to create a context-sensitive dynamic connectivity map through time. Unlike the bipolar networks, activity can be perturbed in a single processing step to move the network into a different region of attractor space.  However, the attractors have approximately equal switching activity.  As the impact of a single switch on the network can be dramatic, potentially disparate regions of the attractor space can be very quickly traversed and explored --- the types of network topology (and hence variation in behaviours) that can be reached in a single processing step is much larger in the unipolar case.  This is perhaps most easily shown in the time to solve each task, which is always statistically faster than the other two network types.  Figure~\ref{figure10}(b) shows a smoother switching frequency for unipolar networks --- as a single switch can have a large impact on the network, no rapid spikes in switching activity are required.  On a hardware level, the bipolar networks can be seen to place a lot of stress on a few key synapses, wheres the unipolar networks rely on moderate switching activity of many synapses --- potentially a more long-lasting strategy.

Stable activity is achieved through synchronised switching and balanced use of both LRS and HRS to moderate the percolation of activity through the network.  As with the bipolar networks, unipolar networks use ``trigger'' neurons to change the action when required.  In this case, the trigger neuron affects a change on a network-wide scale, rather than between only a small subnetwork.  Self-stabilising activity is observed, e.g. the network moves into a state which adds activity which keeps the attractor in the given state even when the neuron that causes the state change is subsequently non-firing.

Unipolar attractors involve more of the network to create the weight oscillator. It is thought that attractor participation is higher as the synapses can cause much more activity peturbation, larger attractors therefore a method of stabilising the network by distributing/replicating certain required internal behaviour so that it is not detrimentally disrupted.  Bipolar networks attractors are generally smaller in terms of participation, but with many more possible states in them.

\subsection{Discussion}

The focus of this paper was the unipolar memristor synapse, which is evolved for two robotics tasks.  Overall, the unipolar memristor is shown to provide numerous performance benefits compared to the two other synapse types in terms of faster attainment of the required behaviour and better generated paths.  The use of two experimental setups (one reactive, one dynamic) allows us to talk with some generality about the results: both the memory ability and reactive behaviour are in some way improved through use of the unipolar memristor.  It has been clearly demonstrated that the more coarse-grained attractors and restricted binary plasticity scheme do not overtly impede the unipolar networks ability to generate highly-fit pathfinding behaviour.

Evolution is shown to find solutions statistically faster on both tasks when the networks used unipolar memristors when compared to the other synapse types.  More expedient goal-finding behaviour can be attributed to the simpler unipolar attractor space compared to bipolar memristors.  Experiments indicate that the ability to perform online behaviour adaptation gives the unipolar memristor a similar advantage over constant synapses. We can view the unipolar memristor as sitting in a ``sweet spot'' in terms of complexity/evolvability (bipolar memristors are capable of richer behaviours but are more difficult to evolve, constant connections are simpler but lack online adaptability).

One lingering question relates to the scalability of the unipolar synapse to more complex tasks.  Due to the simpler switching characteristics and (at least) equivalent performance when compared to the more traditional bipolar synapse, we do not expect scalability to be an issue with one synapse type any more than it would be with the other.  In fact, due to the reduced attractor space of the unipolar network, one may expect it to scale better than the bipolar networks.  Due to their simpler two-state behaviour, unipolar memristors are much easier to repeatedly fabricate {\em en masse} than the more ubiquitous bipolar memristors due to having switching profiles that are less sensitive to synthesis conditions as they are only required to switch between two states, and encapsulate more robust behaviours as they do not rely on potentially complex interactions of specific analog STDP curves to function.  When this knowledge is combined with the experimental results presented herein, interesting questions are raised regarding the direction of neuromorphic engineering, in particular the issue of biological realism vs. computational efficiency.  Do we aim for computing like a human brain, or would some hybrid analog/digital approach be more efficient?  This article provides some basis to the view that a non-biologically realistic, non-Hebbian architecture may present a more easily-traversable path towards neuromorphic computing.

\bibliographystyle{splncs03}
\bibliography{ccos-acalci}
\vspace{12pt}
\end{document}